\documentclass[runningheads]{llncs}

\usepackage{eccv}

\usepackage{eccvabbrv}

\usepackage{graphicx}
\usepackage{booktabs}
\usepackage{subcaption}

\usepackage[accsupp]{axessibility}  

\usepackage{hyperref}

\usepackage{orcidlink}

\usepackage[dvipsnames]{xcolor}

\newcommand{\mn}[1]{{#1}}

\usepackage{times}
\usepackage{epsfig}
\usepackage{graphicx}
\usepackage{amsmath}
\usepackage{amssymb}
\usepackage{pifont}
\usepackage{dsfont}
\usepackage{multirow}
\usepackage{tablefootnote}

\usepackage{subcaption}
\usepackage{wrapfig}

\newcommand{\lsim}{\mathcal{L}_{\text{sim}}}
\newcommand{\lreg}{\mathcal{L}_{\text{reg}}}

\begin{document}

\title{\texttt{NePhi}: Neural Deformation Fields \\ for Approximately Diffeomorphic \\ Medical Image Registration}

\titlerunning{\texttt{NePhi}: Neural Deformation Fields for Medical Image Registration}
%
\author{Lin Tian\inst{1}\orcidlink{0000-0003-0908-5998} \and
Hastings Greer\inst{1} \and
Ra\'ul San Jos\'e Est\'epar\inst{2} \and \\
Roni Sengupta\inst{1} \and
Marc Niethammer\inst{1}}
%
\authorrunning{L.~Tian et al.}
%
\institute{UNC Chapel Hill \and
Harvard Medical School}

\maketitle

\begin{abstract}
This work proposes \texttt{NePhi}, a generalizable neural deformation model which results in approximately diffeomorphic transformations. In contrast to the predominant voxel-based transformation fields used in learning-based registration approaches, \texttt{NePhi} represents deformations functionally, leading to great flexibility within the design space of memory consumption during training and inference, inference time, registration accuracy, as well as transformation regularity. Specifically, \texttt{NePhi} 1) requires less memory compared to voxel-based learning approaches, 2) improves inference speed by predicting latent codes, compared to current existing neural deformation based registration approaches that \emph{only} rely on optimization, 3) improves accuracy via instance optimization, and 4) shows excellent deformation regularity which is highly desirable for medical image registration. We demonstrate the performance of \texttt{NePhi} on a 2D synthetic dataset as well as for real 3D medical image datasets (e.g., lungs and brains). Our results show that \texttt{NePhi} can match the accuracy of voxel-based representations in a single-resolution registration setting. For multi-resolution registration, our method matches the accuracy of current SOTA learning-based registration approaches with instance optimization while reducing memory requirements by a factor of five. Our code is available at \url{https://github.com/uncbiag/NePhi}.

\end{abstract}

\section{Introduction}

\begin{figure}[tp]
    \centering
    \includegraphics[width=\linewidth]{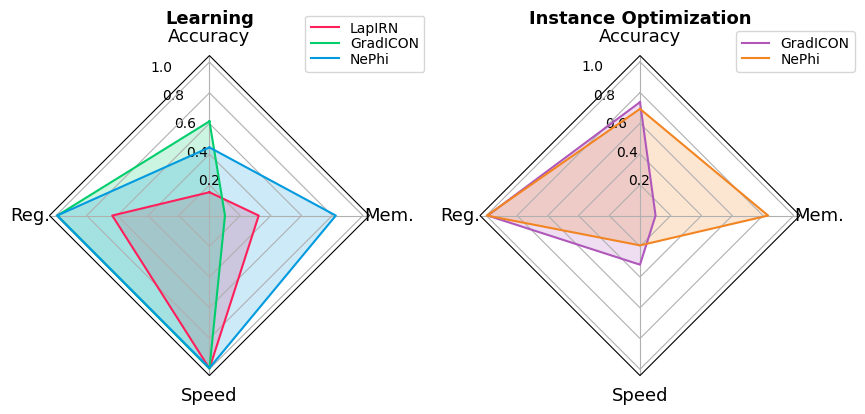}
    \caption{Comparison between \texttt{NePhi} and competing methods based on four design considerations: registration accuracy, regularity of the predicted deformation field, peak memory consumption during training, and inference run-time. \texttt{NePhi} performs slightly worse for pure prediction but on-par with GradICON when combined with instance optimization. In both cases, \texttt{NePhi} shows better deformation regularity and memory consumption compared to other methods. Values are normalized to [0,1] for visualization. Refer to Supp.~\ref{sec:lung_registration_continue} for details.}
    \label{fig:lung_registration_methods_radar}
\end{figure}

 Image registration is a key task in medical image analysis~\cite{crum2004non,sotiras2013deformable}. The goal of image registration is to establish spatial correspondences between image pairs. The ideal registration algorithm should require minimal memory, should be fast and accurate, and should provide plausible transformations. However, while many registration algorithms have been developed over the last several decades~\cite{heinrich2013mrf,vishnevskiy2016isotropic,balakrishnan2018unsupervised,yang2017quicksilver,MokC20,tian2022gradicon,chen2022transmorph}, they typically only provide a subset of these desired qualities. As a consequence, different registration tasks will require different registration algorithms. The goal of this work is to provide \texttt{NePhi}, a neural registration algorithm that allows for significant freedom to adjust registration algorithm properties within the design space of memory consumption, speed, accuracy, and deformation regularity as shown in Fig.~\ref{fig:lung_registration_methods_radar}.  

\noindent{\bf Memory consumption:} Classical optimization-based registration algorithms~\cite{Ashburner07,HeinrichJBS12,heinrich2013mrf,SunNK14,HeinrichPSH14,vishnevskiy2016isotropic,SiebertHH21} that optimize directly over the parameters of a transformation model (e.g., a 3D displacement field) are highly memory efficient. However, current registration research is focused on registration approaches based on deep learning due to achievable improvements in registration accuracy and speed~\cite{yang2017quicksilver,ShenHXN19,DalcaBGS19,BalakrishnanZSG19, MokC20,CVPRMokC20,greer2021icon,tian2022gradicon,chen2022transmorph}. While directly predicting a voxel-represented displacement field with a deep network is fast, memory consumption scales with the number of voxels and can therefore quickly become prohibitive for large 3D image volumes. This then necessitates the use of downsampled images and small batch sizes during training even on modern high-memory GPUs~\cite{MokC20,greer2021icon,chen2022transmorph,tian2022gradicon}. Further, to achieve best performance these approaches perform instance-optimization~\cite{tian2022gradicon} at test time where the neural network parameters are optimized for a given image pair. Hence, memory requirements are high during training \emph{and} testing making these approaches unsuitable for low-memory GPUs, high-resolution images, or high-dimensional vector-valued images. Hence, exploring alternative transformation parameterizations is desirable \mn{to reduce memory consumption; this is expected to become increasingly important, e.g., for very high resolution volumetric microscopy images.} \emph{Our \texttt{NePhi} approach allows for low memory consumption during training and testing as it is based on an implicit neural representation parameterized by a multi-layer perceptron (MLP). Further, it is expected that such a representation can be parameter-efficient by exploiting the expectation that deformation spaces for real data are low-dimensional. In contrast, direct parameterization of a displacement field is very high dimensional; we would for example have 3 million degrees of freedom for a $100^3$ image in 3D.}  

\noindent{\bf Inference speed and accuracy:} Voxel-based deep network approaches are fast but only achieve high accuracies with memory-intensive instance optimization~\cite{tian2022gradicon}. Implicit neural representations of deformations~\cite{han2023diffeomorphic,van2023deformable, wolterink2022implicit,sun2022mirnf,zou2023homeomorphic,wu2023neurepdiff,beg2005computing,van2023robust}, which are parameterized by MLPs, can achieve high accuracy, but are slow as they optimize MLPs from scratch for each image pair. \emph{Our approach allows balancing inference time and accuracy because it predicts global and latent codes on which the MLP is conditioned. Hence a registration solution can be obtained by pure prediction or, if additional inference time is acceptable, can be refined for additional accuracy based on instance optimization.}

\noindent{\bf Regularity:} In medical image registration, transformation regularity is often important and achieved by adding an appropriate regularizer, for example, based on concepts of diffusion~\cite{thirion1998image}, curvature~\cite{holden2007review}, or models from continuum mechanics~\cite{christensen19943d,christensen1996deformable,beg2005computing,vercauteren2008symmetric}. \emph{Our approach provides highly regular transformations by using gradient inverse consistency regularization, a regularizer which has shown to be highly effective for voxel-based deep registration models~\cite{tian2022gradicon}. We further show that using this regularizer only on a very sparse set of locations is sufficient to regularize \texttt{NePhi} thereby further improving memory efficiency.}

\noindent{\bf The contributions of our work are as follows:}
\begin{itemize}
    \item[1)] We propose a generalizable neural deformation field approach, \texttt{NePhi}, which is based on an implicit neural representation via an MLP. In contrast to existing neural deformation models for optimization-based image registration~\cite{han2023diffeomorphic,van2023deformable, wolterink2022implicit,sun2022mirnf,zou2023homeomorphic,wu2023neurepdiff}, \texttt{NePhi} conditions the MLP on global and local codes thereby allowing to \emph{predict} transformations.
    \item[2)] We demonstrate that \texttt{NePhi} can provide high accuracy registration results at significantly lower memory cost than voxel-based deep registration networks.
    \item[3)] While our base \texttt{NePhi} model directly predicts a deformation field, we also provide a hybrid multi-step approach for improved performance. The first step of this multi-step approach uses a low-resolution voxel-based model, which can be computed efficiently thereby retaining the overall efficiency of our approach.
    \item[4)] By integrating a gradient inverse consistency (GradICON) loss into \texttt{NePhi} we obtain approximately diffeomorphic transformations (i.e., transformations that are smooth, have a smooth inverse, and are bijective) even when only evaluating this loss at a sparse set of spatial locations. \mn{Going beyond the existing GradICON approach we propose approximately diffeomorphic neural deformation fields (NDF) which represent the forward and backward deformations with \emph{one} set of latent codes that can be predicted by a CNN in one forward pass.}
    \item[5)] We demonstrate the performance of \texttt{NePhi} on synthetic data, a real 3D lung registration between inhale and exhale computed tomography (CT) images, and a real brain MRI dataset. We evaluate our method across our four critical design considerations (memory consumption, accuracy, regularity, and training and inference time). This evaluation shows that our approach i) is very memory efficient, ii) can achieve excellent registration performance when combined with instance optimization or alternatively quickly predict registrations at reduced accuracy; and iii) results in highly regular transformations. 
\end{itemize}

\section{Related work}
\subsection{Medical Image Registration}\label{sec:related_work_medical_registration}
Image registration is classically formulated as an optimization problem~\cite{Ashburner07,SunNK14,SiebertHH21,HeinrichJBS12,HeinrichPSH14} with respect to a pair of images. The goal is to estimate optimal transformation parameters that minimize a compromise between transformation regularity and the dissimilarity between the transformed moving image with respect to the fixed image. As generally no closed-form solutions to these optimization problems exist they are solved using iterative numerical optimization (typically using a gradient descent scheme), resulting in significant computation time. To improve capture range multi-resolution approaches have been proposed~\cite{ants2008,MAES1999373}. To reduce registration run-time, learning-based methods~\cite{yang2017quicksilver,ShenHXN19,DalcaBGS19,BalakrishnanZSG19, MokC20,CVPRMokC20,greer2021icon,tian2022gradicon} have been developed which replace numerical optimization by regression via a deep neural network at inference time. The pioneering works~\cite{yang2017quicksilver,DalcaBGS19} demonstrated how convolutional neural networks (CNN) can be used for learning-based registration. Subsequent developments added multi-step and multi-resolution capabilities~\cite{MokC20,CVPRMokC20,greer2021icon,tian2022gradicon} to improve registration accuracy; these approaches achieve competitive accuracies to optimization-based multi-resolution registration methods but with significantly less run-time. Instance-optimization, where the registration networks are finetuned for a given image pair, can further increase registration performance~\cite{tian2022gradicon}. While learning-based multi-resolution registration methods provide state-of-the-art performance and nearly instantaneous inference, they consume significant memory during training and instance optimization (IO), because the used CNNs predict dense deformation fields and feature maps which becomes especially costly in 3D. Registrations at high spatial resolutions for 3D image volumes are therefore not easily possible. Further, these approaches are relatively slow when using IO as large neural networks are finetuned. \emph{Our motivation is therefore to develop \texttt{NePhi}, a neural deformation model which has a better memory profile and allows balancing between runtime, memory consumption, and registration accuracy.} 

\subsection{Neural Deformation Models}
Functional representations of deformation fields (parameterized via MLPs) for natural images in dynamic scenes~\cite{gao2021dynamic,liu2022devrf, liu2022devrf, park2021hypernerf,pumarola2021d,tretschk2021non,park2021nerfies,li2021neural}, for dynamic objects~\cite{lei2022cadex, sundararaman2022reduced,wang2022generative,duggal2022topologically,niemeyer2019occupancy}, and for animatable humans~\cite{xu2021h, zhang2022ndf,zheng2022avatar, chen2021snarf,grassal2022neural,liu2021neural,peng2021animatable,shao2022doublefield} have attracted significant recent interest. These approaches focus on reconstructing an underlying 3D scene, object, or human while accounting for 
 existing motion between acquired 2D images. However, only limited work exists on using such functional representations for medical image registration~\cite{han2023diffeomorphic,van2023deformable, wolterink2022implicit,sun2022mirnf,zou2023homeomorphic,wu2023neurepdiff,van2023robust}. Existing approaches either use an MLP-parameterized function to directly represent a displacement vector field (DVF)~\cite{zou2023homeomorphic,wolterink2022implicit} or to represent velocity or momentum fields~\cite{han2023diffeomorphic,sun2022mirnf,wu2023neurepdiff} to capture large deformations. In velocity or momentum field approaches a transformation is obtained by numerical integration; this indirect parameterization ensures diffeomorphic transformations. but is computationally costly because of the required numerical integrations. \emph{Most importantly, all existing registration work based on neural deformation models only focuses on \emph{optimization-based} registration, which is slow compared to learning-based registration which predicts transformations at test time. In contrast, our proposed \texttt{NePhi} approach is suitable for optimization-based \emph{and} learning-based registration.}

\section{Background}
Given two images $I^A:\Omega^A\,{\to}\,\mathds{R}$ and 
$I^B:\,\Omega^B{\to}\,\mathds{R}$, where $\Omega^{A,B}\in\mathds{R}^d$ indicates the domain of the images, our goal is to find a mapping $\varphi^{AB}: \mathds{R}^d\to\mathds{R}^d$ such that $I^A \circ \varphi^{AB} \approx I^B$\footnote{A perfect match can in general not be achieved due to image differences or image noise.}. This continuous transformation map, $\varphi^{AB}$, can be parameterized as a function with parameters $\tau$, denoted as $\varphi^{AB}(x;\tau)$, where the parameters depend on the function class to be parameterized. E.g., these could be parameters for an affine transform $\varphi^{AB}(x;\tau) = Tx+b$ where $\tau=\{T,b\}$, a discretized vector field, $\tau$, on a voxel grid $\varphi^{AB}(x;\tau) = x+$ interpolate($\tau$ at $x$), or the condition latent code $z$ to an MLP $\varphi^{AB}(x;\tau)=\varphi_{\theta}(x;z)$ with $\tau=z$ and $\varphi_{\theta}$ being the MLP.

\textbf{Learning-based registration.} Here, one trains a neural network (NN) $f_{\theta}(I^A_i, I^B_i) = \tau$ that outputs the parameters of a chosen transformation model $\varphi$. We denote the predicted transformation field as $\varphi(x;\tau)$. The NN is trained over a set of image pairs $I=\{(I^A_i, I^B_i)\}_{i=1}^N$ by solving 
\begin{equation}
\theta^*{=}\underset{\theta}{\rm argmin} \frac{1}{N}\sum_{i=1}^N \lsim\left(I^A_i\circ{\varphi(x;\tau)},I^B_i\right)\,{+}\,\lambda\lreg(\varphi(x;\tau))\,, \tau=f_{\theta}(I^A_i, I^B_i)\,,
\label{eq:learning_loss_voxel}
\end{equation} 
where $\lsim(\cdot,\cdot)$ is the \emph{similarity measure} and $\lreg(\cdot)$ is a \emph{regularizer} with $\lambda\geq 0$. 

\textbf{Diffeomorphic transformations.} In medical image registration it is frequently desirable to obtain diffeomorphic transformation maps, which are desirable because they allow mapping smoothly between image spaces (e.g., between an image and an atlas space) without worrying about transformation folds. As discussed above, the standard approach to obtain diffeomorphic transformations by construction is by parameterizing them by velocity~\cite{vercauteren2008symmetric} or momentum fields~\cite{beg2005computing}. Instead, we will use a regularizer (see Sec.~\ref{sec:method_neural_deformation_fields}) that directly encourages invertibility and empirically results in approximately diffeomorphic transformations.


\begin{figure*}[tp]
    \centering
    \includegraphics[width=0.95\linewidth, trim={0 0cm 0 0cm}, clip]{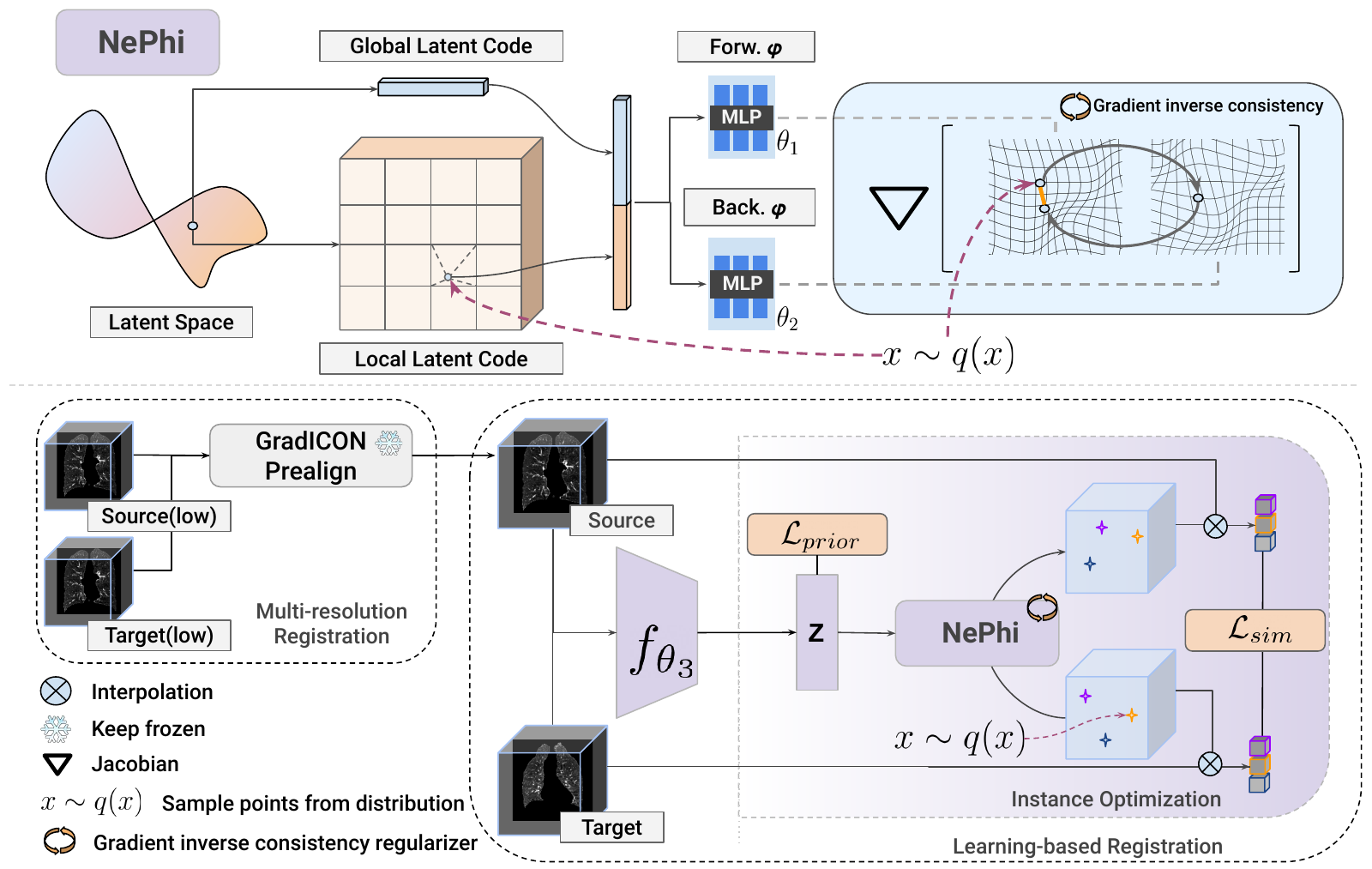}
    \caption{The framework of \texttt{NePhi}. Top: the design of the generalizable neural deformation fields with gradient inverse consistency regularizer (Sec.~\ref{sec:method_neural_deformation_fields} and Sec.~\ref{sec:method_hybrid_condition}). Bottom: the overall framework of how to train neural networks to predict \texttt{NePhi} (Sec.~\ref{sec:method_training}).}
    \label{fig:nephi_pipeline}
\end{figure*}

\section{Methods}
\label{sec:methods}

This section describes our approximately diffeomorphic neural deformation fields (NDFs) (Sec.~\ref{sec:method_neural_deformation_fields}), how we obtain generalizable neural deformation fields  (Sec.~\ref{sec:method_hybrid_condition}), and lastly our training approach (Sec.~\ref{sec:method_training}).  

\subsection{Approximately Diffeomorphic  NDFs}\label{sec:method_neural_deformation_fields}

We build on the approach of Tian et al.~\cite{tian2022gradicon} to obtain \emph{approximately} diffeomorphic neural transformation maps. This approach regularizes transformations based on a gradient inverse consistency loss to encourage diffeomorphic transforms. Specifically, given a forward transformation map $\varphi^{AB} = \text{Id}+f_\theta(I^A, I^B)$ and a backward map $\varphi^{BA}=\text{Id}+f_\theta(I^B, I^A)$, where $\text{Id}$ denotes the identity map, regularity can be encouraged by minimizing the loss
\begin{equation}
\mathcal{L}_{reg}=\left\|\nabla\left[\varphi^{AB}\circ\varphi^{BA}\right]-\mathbf{I} \right\|_F^2\,,
\end{equation}
where $\mathbf{I}$ is the identity matrix and $\nabla$ denotes the Jacobian. To combine this loss with our neural representation of \texttt{NePhi}, we use \emph{two} MLPs, denoted  $\varphi_{\theta_1}$ and $\varphi_{\theta_2}$, to represent the continuous forward transformation $\varphi^{AB}$ and backward transformation $\varphi^{BA}$, respectively. To make these MLPs generalizable we condition them on the \emph{same} latent code $z$, which is possible because of our design choice to use two MLPs. With such a design, we can then express the gradient inverse consistency loss as 
\begin{equation}
\mathcal{L}_{reg}(\theta_1,\theta_2,z) 
    =E_{x\sim{q(x)}}\left\|\nabla\left[\varphi_{\theta_1}(\varphi_{\theta_2}(x;z);z) - x\right]\right\|_F^2\,,
\end{equation}
where $x$ denotes a point coordinate which we uniformly sample from distribution $q(x)$ across the image domain.

\subsection{Hybrid Conditioning}\label{sec:method_hybrid_condition}
To allow \texttt{NePhi} to generalize to new transformations (for new image pairs) we use the latent code, $z$, introduced above to condition $\varphi_{\theta_1}$ and $\varphi_{\theta_2}$. Specifically, $z$ is composed of a global latent code \emph{vector} $z_g\in\mathbb{R}^{C\times 1}$ and a local latent code \emph{map} $z_l\in\mathbb{R}^{C\times H \times W}$ for a 2D transformation and $z_l\in\mathbb{R}^{C\times D \times H \times W}$ for a 3D transformation. Note that the local latent code is represented at a lower resolution than the input images. For any point $x$ we then obtain a local latent code via bilinear (trilinear in 3D) interpolation of the local latent codes. We denote the interpolated local latent code at point $x$ by $z_l(x)\in\mathbb{R}^{C\times 1}$. For a given $x$, the overall latent code, $z\in\mathbb{R}^{2C\times 1}$ is then obtained by  concatenation
\begin{equation}
    z(x) = [z_g^T, z_l(x)^T]^T\,.
\end{equation}
We use the following loss from~\cite{park2019deepsdf} to encourage a compact latent space
\begin{equation}
    \mathcal{L}_{prior}(z_g,z_l) = \frac{1}{\sigma^2}||z_g||^2_2 + \frac{1}{|\Omega_{z_l}|}\sum_{x\in\Omega_{z_l}}\frac{1}{\sigma^2}||z_l(x)||^2_2\,
\end{equation}
where $\Omega_{z_l}$ represents the grids where the $z_l$ are defined.

\subsection{Training}\label{sec:method_training}
\noindent
Here, we describe how \texttt{NePhi} can be used for learning-based registration. Fig.~\ref{fig:nephi_pipeline} (bottom) illustrates the approach. Supp.~\ref{sec:implementation_details} provides implementation details.

\textbf{Learning-based \texttt{NePhi}.}
We condition \texttt{NePhi} on latent codes because we want \texttt{NePhi} to provide equivalent functionality to the voxel representation used in learning-based registration. In a learning-based registration network, a CNN directly predicts the transformation given a pair of images during inference. In \texttt{NePhi}, we achieve a similar functionality by letting a neural network $f_{\theta_3}$ predict the latent codes $z_g$ and $z_l$ based on an input image pair. Given a training dataset $I=\{(I^A_i, I^B_i)\}_{i=1}^N$, we train the neural network $f_{\theta_3}$ and NDFs $\varphi_{\theta_1}$, $\varphi_{\theta_2}$ by minimizing 
\begin{multline}
\{\theta_1,\theta_2,\theta_3\}=\underset{\theta_1,\theta_2,\theta_3}{\rm argmin}~\frac{1}{N}\sum_{i=1}^N\lsim\left(I^A_i\circ\varphi_{\theta_1}(\cdot;z),I^B_i\right) +\lsim\left(I^A_i,I^B_i\circ\varphi_{\theta_2}(\cdot;z)\right)\\+\,\lambda_1\lreg(\theta_1,\theta_2,z)+\lambda_2\mathcal{L}_{prior}(z)\,, z=f_{\theta_3}(I^A_i,I^B_i)\,,
\label{eq:learning_nephi}
\end{multline} 
where $\mathcal{L}_{sim}$ is the similarity loss: in this work, we use normalized cross correlation (NCC). As $\varphi_{\theta_1}$ and $\varphi_{\theta_2}$ are continuous functions that can represent transformations at any spatial location, we randomly sample two different sets of points within the image space to compute $\lsim$ and $\lreg$. 

\textbf{\texttt{NePhi} Instance Optimization.}
Starting from the predicted latent code from the learning-based \texttt{NePhi}, we optimize over the latent code and the parameters of $\varphi_{\theta_1}$ and $\varphi_{\theta_2}$ by minimizing the following loss at test time when given one pair of images: 
\begin{multline}
\{z,\theta_1,\theta_2\}=\underset{z,\theta_1,\theta_2}{\rm argmin}~\lsim\left(I^A\circ\varphi_{\theta_1}(\cdot;z),I^B\right)+\lsim\left(I^A,I^B\circ\varphi_{\theta_2}(\cdot;z)\right)\\+\,\lambda_1\lreg(\theta_1,\theta_2,z)+\lambda_2\mathcal{L}_{prior}(z)\,. 
\label{eq:optimization_nephi}
\end{multline} 
\emph{During training the latent code is predicted by $f_{\theta_3}$. During instance optimization, the latent code is optimized over.}

\textbf{Multi-resolution \texttt{NePhi}}
To improve inference performance, we propose a multi-resolution approach based on \texttt{NePhi}. We combine a low-resolution voxel-based registration CNN with a high-resolution learning-based \textbf{NePhi} as shown in Figure~\ref{fig:nephi_pipeline}. This approach retains the good memory efficiency of \texttt{NePhi} as the voxel-based registration at low-resolution is also fast and memory efficient. See Fig.~\ref{fig:exp_memory_usage_varying_resolution} for an illustration of \texttt{NePhi}'s memory efficiency for high resolution images. Specifically, we train a voxel-based registration neural network on low-resolution images and keep it frozen while training a subsequent high-resolution registration step based on \texttt{NePhi}.

\textbf{Sampling Strategy}
As we will observe in Sec.~\ref{sec:exp_ablation_number_of_sampling_points}, regular transforms can be obtained using very few sampled points for the regularizer. Thus, we sample two different sets of points uniformly within the image domain $[-1,1]$ to compute $\lsim$ and $\lreg$, with a much smaller number of points for $\lreg$ than for $\lsim$.

\section{Experiments}

\subsection{Datasets}
We use the following datasets for our experiments.

\noindent
\textbf{Triangles and Circles (T\&C)} is a synthetic 2D dataset~\cite{greer2021icon} where images are either solid or hollow triangles or circles. We randomly generate a training set of 10000 pairs of images of size $512\times512$ and a validation set with 1000 pairs of images of the same size for solid and for hollow shapes. We use this dataset for an ablation study.

\noindent
\textbf{COPDGene \normalfont{\cite{regan2011genetic}}.} We use a subset of 999 inspiration/expiration lung computed tomography (CT) image pairs from the {\bf COPDGene}
study\footnote{\url{https://www.ncbi.nlm.nih.gov/gap}}~\cite{regan2011genetic} with provided lung segmentation masks for training. We resample the images to an isotropic spacing of 2~mm, leading to a $175{\times}{175}{\times}{175}$ image grid. CT intensities correspond to Hounsfield units. We clamp them within $[-1000,0]$ and scale them linearly to $[0,1]$. We then apply the lung segmentation mask to the images to extract the lung region of interest (ROI). We use 899 pairs for training and 100 pairs for validation. 

\noindent
\textbf{DirLab\normalfont{\cite{castillo2013reference}}.} This dataset contains 10 pairs of inspiration/expiration lung CT images (from COPDGene, different from the 999 pairs above) with 300 anatomical landmarks per pair, manually identified by an expert in thoracic imaging. We process these images in the same way as the other \textbf{COPDGene} images. This dataset is used for testing.

\noindent
\mn{\textbf{HCP\normalfont{\cite{van2012human}}. We use a subset of the young adult T1-weighted brain images of the Human Connectome Project (HCP) for training. We use 28 manually segmented subcortical regions for evaluation~\cite{rushmore2022anatomically,rushmore_r_jarrett_2022_dataset}. Following~\cite{tian2022gradicon} we use 1076 images for training and 44 for testing.}}

\subsection{Baselines}
We compare \texttt{NePhi} to the following optimization-based method and two learning-based methods.

\textbf{PTVReg}~\cite{vishnevskiy2016isotropic} is a state-of-the-art optimization-based approach on the DirLab dataset. We report the result on DirLab based on the official implementation, where hyper-parameters were obtained by grid-search on the DirLab dataset itself. Results are therefore likely overly optimistic.

\textbf{LapIRN}~\cite{MokC20} is the winner of the Learn2Reg registration challenge. To be consistent with the transformation model of \texttt{NePhi}, we use LapIRN with displacement vector fields (DVFs) from the official implementation and train it with the COPDGene dataset.

\textbf{GradICON}~\cite{tian2022gradicon} is currently the best learning-based registration method on the DirLab dataset. It is a multi-resolution registration neural network composed of three sub-networks. Here a subnetwork  receives as input the warped images from the previous network. We compare to GradICON in two ways: 1) based on the original settings, and 2) by replacing the multi-resolution network architecture with a single-resolution architecture. All other settings are kept the same as described in~\cite{tian2022gradicon}. The goal of the single-resolution comparision is to provide a fair comparison to a single-resolution \texttt{NePhi} model. For GradICON(IO), we use 50 iterations as in the original paper.

\begin{table*}[htp]
    \centering
    \caption{Registration performance on \textbf{DirLab}. Registration methods are categorized as optimization-based (O), multi-resolution learning-based (M), and single-resolution learning-based (S). mTRE is the landmark error. $\%|J|_{<0}$ is the voxel percentage with negative Jacobian determinant for the transformation map, which measures the regularity of the transformation. Rep. shows the number of parameters of a representation and Enc. the number of parameter of the encoder network. The inference time for IO indicates the time required for the optimization iterations. \mn{Our \texttt{NePhi} approach combines good performance with excellent registration regularity while reducing memory consumption significantly in comparison to the fully voxel-based GradICON and LapIRN registration networks.}}
    \label{tab:registration_lung}
    \resizebox{0.9\textwidth}{!}{%
    \begin{tabular}{ccccccccc}\toprule
        \multicolumn{2}{c}{Method} &  IO&mTRE$\downarrow$ & $\%|J|_{<0}\downarrow$ & Rep. & Enc. & Time(Inference) & Peak Mem.(Train)$\downarrow$   \\
        && &(mm)&& (M) & (M) &(s)&(MB)\\
        \multicolumn{2}{c}{Initial} &  &23.36 & \textemdash & \textemdash & \textemdash&\textemdash &\textemdash\\ \midrule
        O & PTVReg~\cite{vishnevskiy2016isotropic}  & & 0.84& 0.6& \textemdash & \textemdash & 442&\textemdash \\
        \midrule
        
         \multirow{4}{*}{S}& GradICON &  &5.41 &  1.4e-4  & 0 & 17.67 & 0.14 & 6394  \\
         & GradICON&  $\checkmark$ &2.10  &  3.5e-4  & 0 & 17.67 &  16.08 & 6394 \\
         & \texttt{NePhi} &  &5.44 & 0.0& 0.28 & 3.07 & 0.46& \textbf{2466}\\
         & \texttt{NePhi} &  $\checkmark$ &1.73 & 0 & 0.28 & 0 & 13.60 & 2330 \\ \midrule
         \multirow{5}{*}{M} & LapIRN &  &4.24 & 1.1e-2 & 0 & 0.92 & 0.24 &  10186\\ 
        & GradICON &  &1.93 & 2.6e-4 & 0 & 70.68 & 0.16 & 13482 \\ 
        & GradICON&  $\checkmark$ &1.31 & 2.6e-4 & 0 & 70.68 & 47.63 & 13482 \\
        & \texttt{NePhi} &  &2.78 & 2.4e-4& 0.28 & 3.07 & 0.52& \textbf{2852}\\
         & \texttt{NePhi}&  $\checkmark$ &1.53 & 1.1e-4 & 0.28 & 0 & 56.42 & 2330 
        \\ \bottomrule
    \end{tabular}
    }
\end{table*}

\subsection{Registration Performance on DIRLab and HCP}
\label{sec:lung_registration}
Table~\ref{tab:registration_lung} compares \texttt{NePhi} to baselines on the DirLab dataset. Note that the iterations of \texttt{NePhi}(IO) are adjusted to match the runtime of GradICON(IO) for a fair comparison. 

For single-resolution registration, our method achieves comparable registration accuracy (5.44 mm) to GradICON (5.41 mm), the analogous baseline for a voxel-represented deformation, while using only \emph{$41.4\%$} of the memory required for both training and inference. With instance optimization, \texttt{NePhi} with 200 iterations reaches better performance than GradICON with 50 iterations (1.73 mm versus 2.10 mm) at a comparable run-time, but again requires less memory. Moreover, the deformation fields predicted by \texttt{NePhi} contain \emph{no} folds, indicating better regularity than GradICON and much better regularity than the optimization-based SOTA method PTVReg.

For multi-resolution registration, \texttt{NePhi} achieves better registration accuracy and regularity than LapIRN. Though the inference result of \texttt{NePhi} is slightly worse than for multi-resolution GradICON, memory consumption is \mn{reduced by $\sim79\%$ compared to}  GradICON. This memory efficiency carries over to instance optimization, where the performance gap between \texttt{NePhi}(IO) and GradICON(IO) is further reduced for the same runtime. \mn{See Fig.~\ref{fig:lung_demo_supp} in the supplementary material for an example registration result.}

\begin{table}[htp]
\caption{\mn{Evaluating the multi-resolution \texttt{NePhi} on the HCP brain dataset. As for the DirLab dataset, \texttt{NePhi} provides competitive registration accuracy with excellent spatial regularity and significantly reduced memory requirements compared to the voxel-based GradICON registration network during training.}}
    \label{exp:evaluation_HCP}
    \centering
    \begin{tabular}{cccc} \toprule
         &  DICE$\uparrow$ &  $\%|J|_{<0}\downarrow$&  PeakMem.(Train)$\downarrow$  \\ \midrule
         Initial & 53.4 & \textemdash & \textemdash  \\
         GradICON&  78.7&  1.2e-3& 6716  \\
         GradICON(IO)& 80.5& 4e-4& 6716 \\
         NePhi&  76.2&  0& \textbf{1760}  \\
         NePhi(IO)& 78.9  & 1.4e-3 & \textbf{1426}  \\ \bottomrule
    \end{tabular}
\end{table}

\mn{We also test \texttt{NePhi} on the HCP brain MRI dataset following the same evaluation procedure as in~\cite{tian2022gradicon}. Tab.~\ref{exp:evaluation_HCP} indicates that \texttt{NePhi} shows competitive performance with significantly reduced memory requirements.}

\subsection{Memory consumption across image resolutions} 
\label{sec:exp_memory_usage_varying_resolution}
In the previous experiment, we demonstrated that \texttt{NePhi} exhibits superior memory efficiency compared to state-of-the-art (SOTA) learning-based registration methods for images of size $175\times{175}\times{175}$. Here, we extend our analysis to show that the memory efficiency of \texttt{NePhi} becomes more pronounced as the input image resolution increases. 

\begin{figure}[ht]
    \centering
    \includegraphics[width=0.6\linewidth]{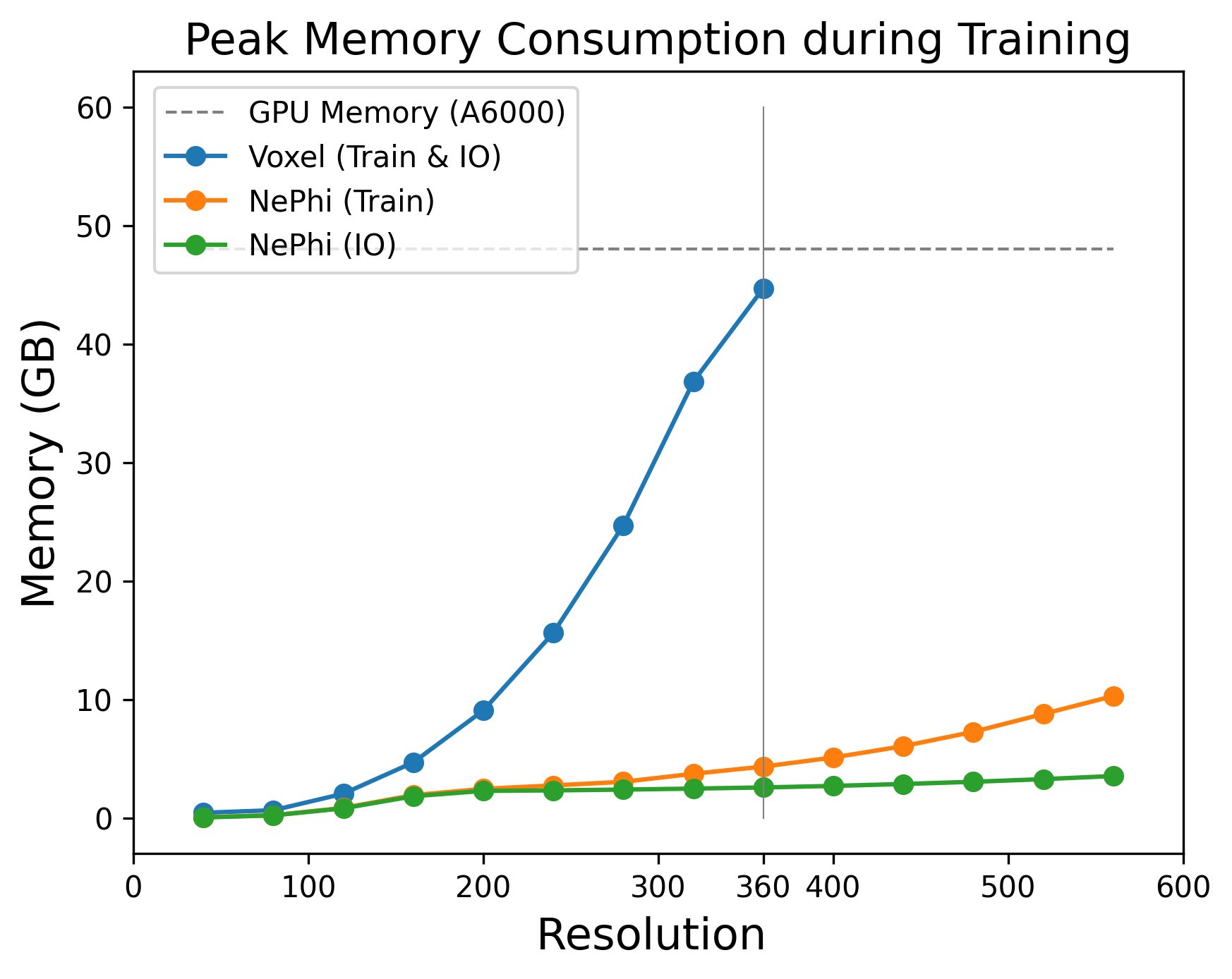}
    \caption{Comparison of peak memory consumption between voxel-representated deformations and \texttt{NePhi} during training and instance optimization (test-time optimization) for 3D images. Details regarding the experiment setting can be found in Sec.~\ref{sec:exp_memory_usage_varying_resolution}. \mn{We observe that memory consumption for voxel-based approaches scales poorly whereas \texttt{NePhi} requires much less memory during training and testing thereby opening opportunities for high-resolution image registration. Such high-resolution capabilities are expected to become increasingly important for example for the registration of very large volumetric miscroscopy images.}}
    \label{fig:exp_memory_usage_varying_resolution}
\end{figure}

Figure~\ref{fig:exp_memory_usage_varying_resolution} compares peak memory use between GradICON and \texttt{NePhi} during training and instance optimization (IO) in the single-resolution setting. The results show that peak memory use escalates rapidly for voxel-represented transformations as the size of the input images increases. \mn{We see that going from $160^3$ to $320^3$, memory use for a voxel--based approach jumps from $\sim 4.8$GB to $\sim37.7$GB with only one pair of images and hits the GPU memory limit for a $48$ GB GPU around $360^3$. In contrast, \texttt{NePhi} shows a significantly slower increase in peak memory use opening the possibility for 3D registrations for significantly larger image volumes}. The peak memory use for \texttt{NePhi} is further reduced because optimization is carried out \emph{only} over the parameters of the neural deformation fields, eliminating the need to optimize over the feature encoder. In contrast, optimizing over voxel-represented transformation fields with gradient inverse consistency as a regularizer (see Supp.~\ref{sec:voxel_vs_nephi_with_gradient_inverse_consistency}), is not feasible. Peak memory consumption during IO remains high for voxel-represented transformation, equaling that of the training phase.

\begin{minipage}{0.45\textwidth}
    \centering
     \captionof{table}{Performance of \texttt{NePhi} on {\bf DirLab} when trained on  different resolutions of the COPDGene dataset. \mn{We observe that regularity is maintained across resolutions while registration performance improves for higher-resolution images.}}
    \label{tab:exp_lung_performance_varying_resolution}
    \resizebox{\textwidth}{!}{
    \begin{tabular}{ccccc} \toprule
         Shape&  mTRE$\downarrow$& $\%|J|_{<0}\downarrow$ & mTRE(IO)$\downarrow$&$\%|J|_{<0}\downarrow$ \\ \midrule
         21&  16.85& 0 & 11.49&0\\
         43&  12.77& 0 & 8.20&0\\
         87&  9.31& 0 & 3.48&0\\
         175&  5.44& 0 & 1.57&3e-5\\
         256&  5.30& 0 & 1.46&3e-4\\ \bottomrule
    \end{tabular}
    }
\end{minipage}\hfill
\begin{minipage}{0.45\textwidth}
    \centering
    \captionof{table}{mTRE (mm) and run-time with varying number of IO iterations \mn{for DirLab}. \texttt{NePhi}(M) outperforms \texttt{NePhi}(S) for real-time inference. Both methods reach satisfying results with reasonable run-time when using IO. \texttt{NePhi}(M) shows better performance for fewer iterations.}
    \label{tab:varying_io_iterations}
    \resizebox{\textwidth}{!}{
    \begin{tabular}{cccccc} \toprule
         IO Iterations &  Initial&  100 &  200 &  800 & 1600 \\ \midrule
         Time(s)&  & 6.76 & 13.60 & 56.42 & 112.68 \\ \midrule
         \texttt{NePhi}(S)&  5.44&  1.94&  1.73&  1.57& 1.51\\
         \texttt{NePhi}(M)&  2.78&  1.78&  1.67&  1.53& 1.49\\ \bottomrule
    \end{tabular}
    }
\end{minipage}

Beyond memory considerations, Table~\ref{tab:exp_lung_performance_varying_resolution} shows the accuracy and regularity properties based on the predicted deformation field for \texttt{NePhi} trained on various image resolutions \mn{for the DirLab dataset}. We observe that regularity is maintained across all resolutions, and landmark errors decrease as expected with increasing resolution. This analysis underscores the positive correlation between high-resolution input images and improved registration performance, emphasizing the necessity of exploring high-resolution registration.

\subsection{Trade-off between time and accuracy}

Table~\ref{tab:varying_io_iterations} shows registration accuracies for instance optimization (IO) using single-resolution \texttt{NePhi} and multi-resolution \texttt{NePhi} for varying numbers of IO iterations \mn{on the DirLab dataset}. We observe that using more iterations results in better registration performance. Multi-resolution registration provides slightly worse than SOTA results while providing fast memory-efficient inference (0.46s for single-resolution \texttt{NePhi} and 0.52s for multi-resolution \texttt{NePhi}). To pursue the best accuracy, either single-resolution or multi-resolution \texttt{NePhi}(IO) can be used with reasonable run-time ($\sim56s$). These results demonstrate that \texttt{NePhi} can effectively balance accuracy and run-time based on registration requirements. 

\subsection{Ablation study}
This section investigates the impact of i) the number of points used for $\lreg$, and ii) the different choices of $\lreg$ when combined with NePhi NDF.

\begin{figure*}[htp]
    \centering
    \begin{subfigure}[t]{0.32\textwidth}
        \centering
        \includegraphics[width=0.92\textwidth]{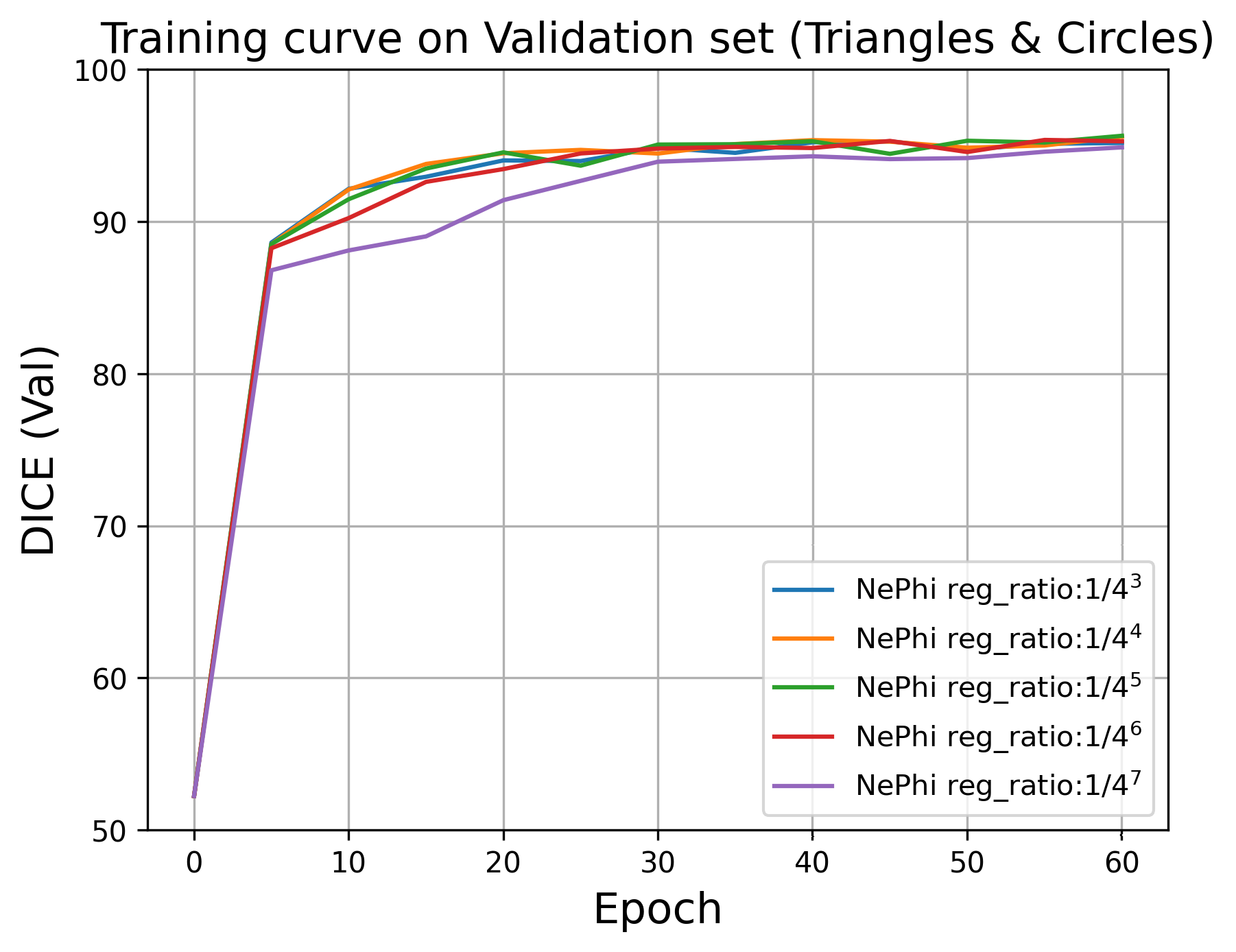}
        \caption{}
        \label{fig:nephi_learning_varying_reg_ratio_loss}
    \end{subfigure}
    \begin{subfigure}[t]{0.32\textwidth}
        \centering
        \includegraphics[width=\textwidth]{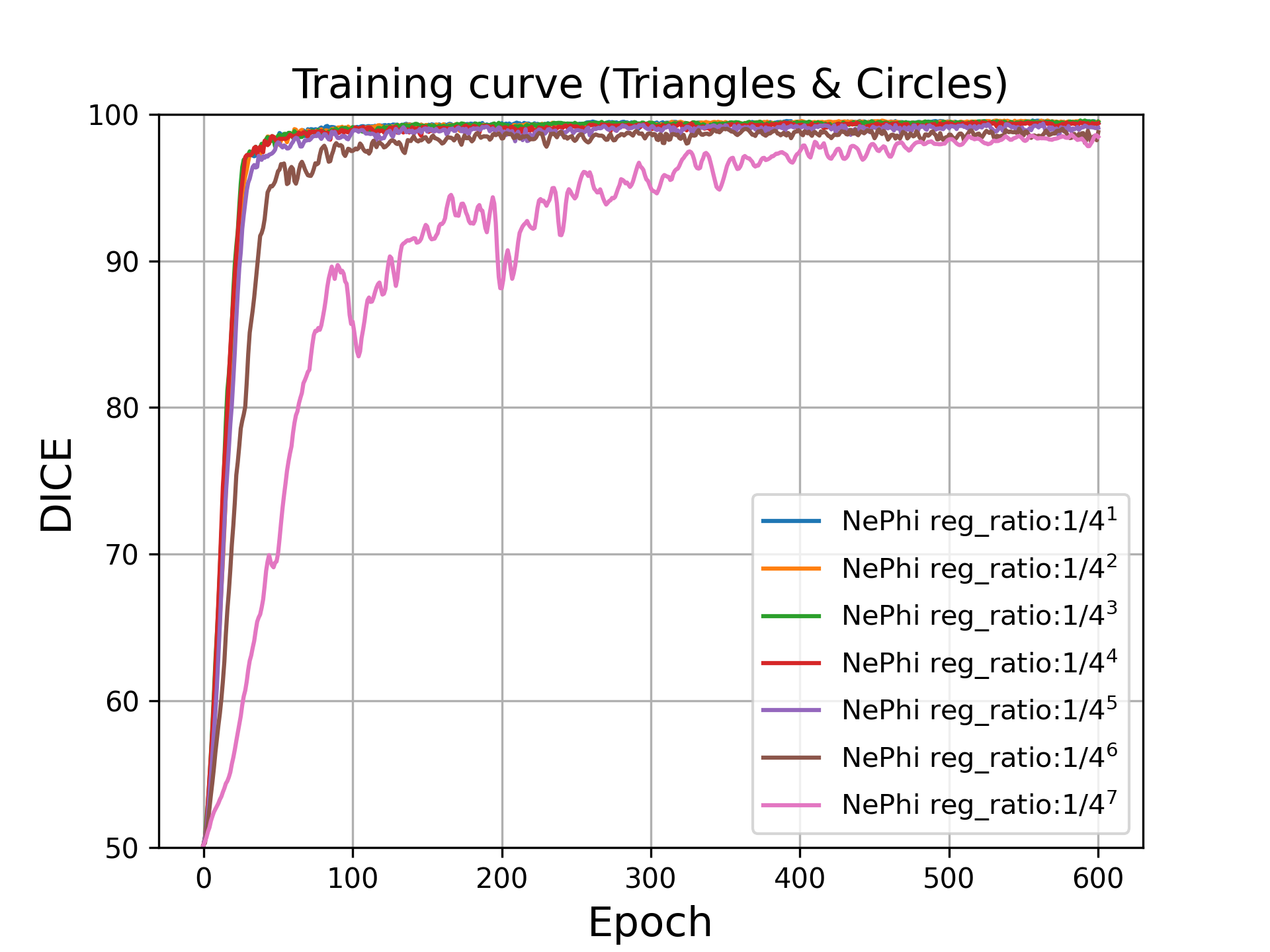}
        \caption{}
        \label{fig:nephi_optimization_varying_reg_ratio_dice}
    \end{subfigure}
    \begin{subfigure}[t]{0.32\textwidth}
        \centering
        \includegraphics[width=\textwidth]{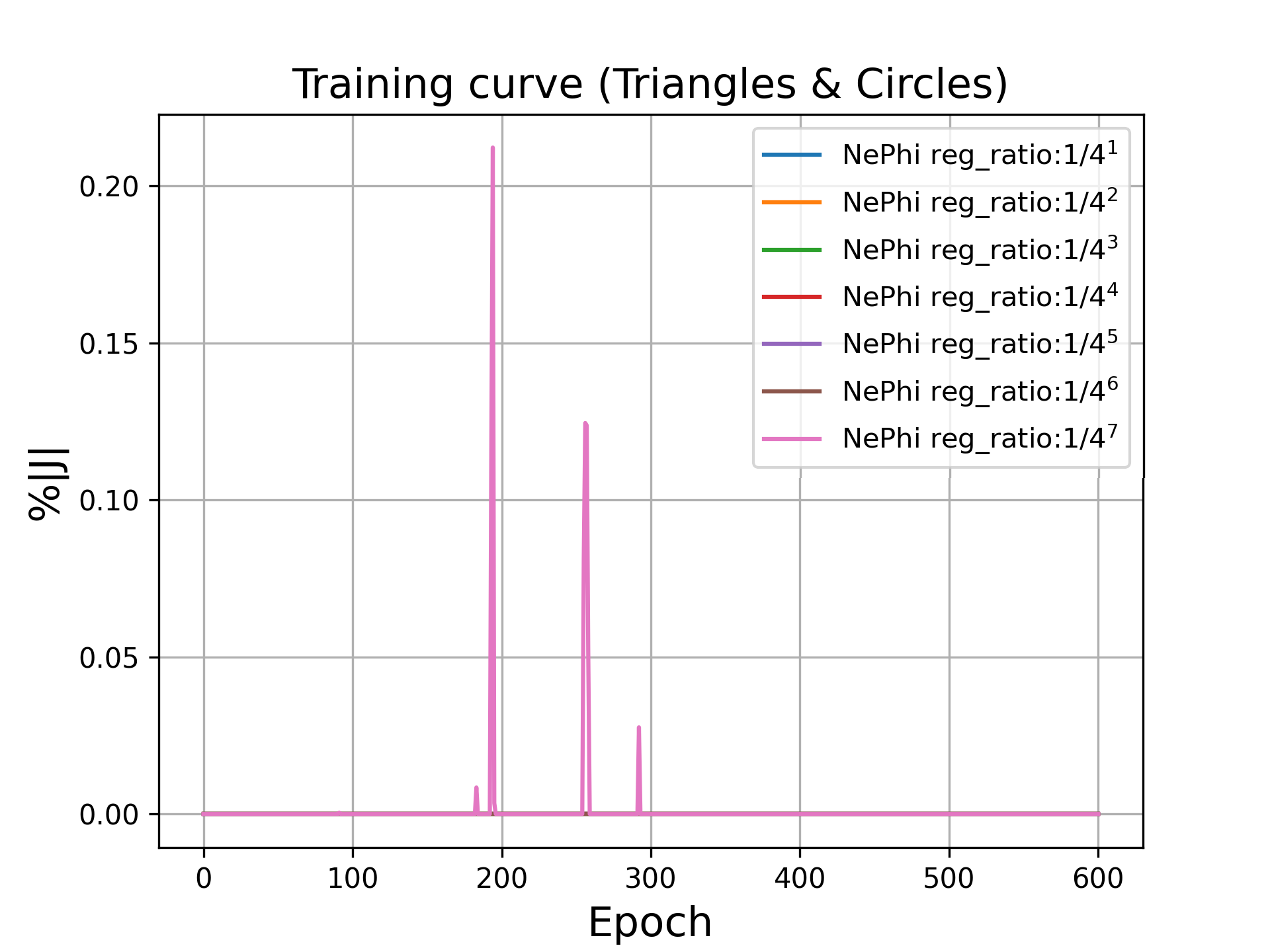}
        \caption{}
        \label{fig:nephi_optimization_varying_reg_ratio_fold}
    \end{subfigure}
    
    \begin{subfigure}[t]{0.32\textwidth}
        \centering
        \includegraphics[width=\textwidth]{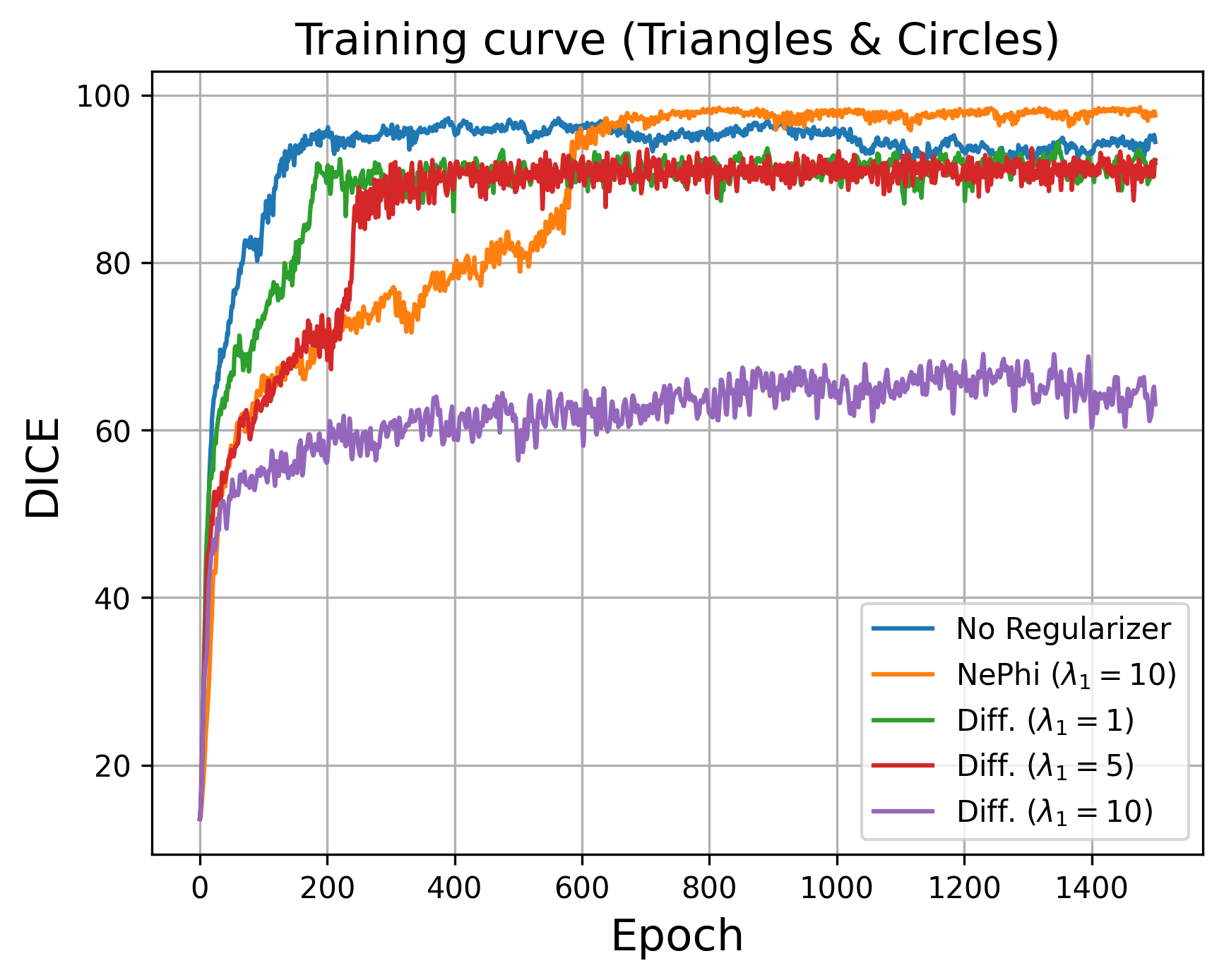}
        \caption{}
        \label{fig:nephi_vs_diffusion_dice_hollow}
    \end{subfigure}
    \begin{subfigure}[t]{0.32\textwidth}
        \centering
        \includegraphics[width=\textwidth]{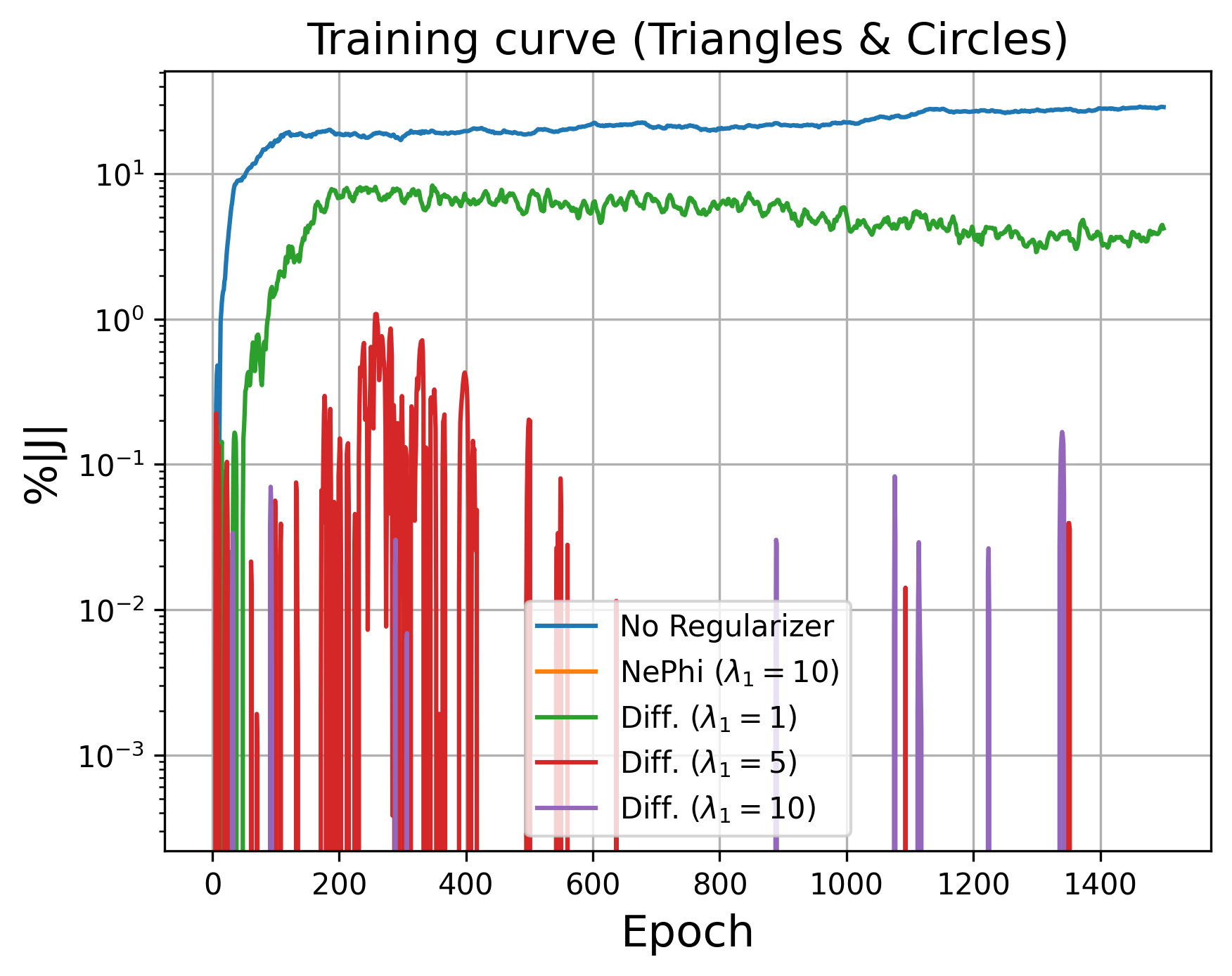}
        \caption{}
        \label{fig:nephi_vs_diffusion_fold_hollow}
    \end{subfigure}
    \begin{subfigure}[t]{0.32\textwidth}
        \centering
        \includegraphics[width=0.9\textwidth]{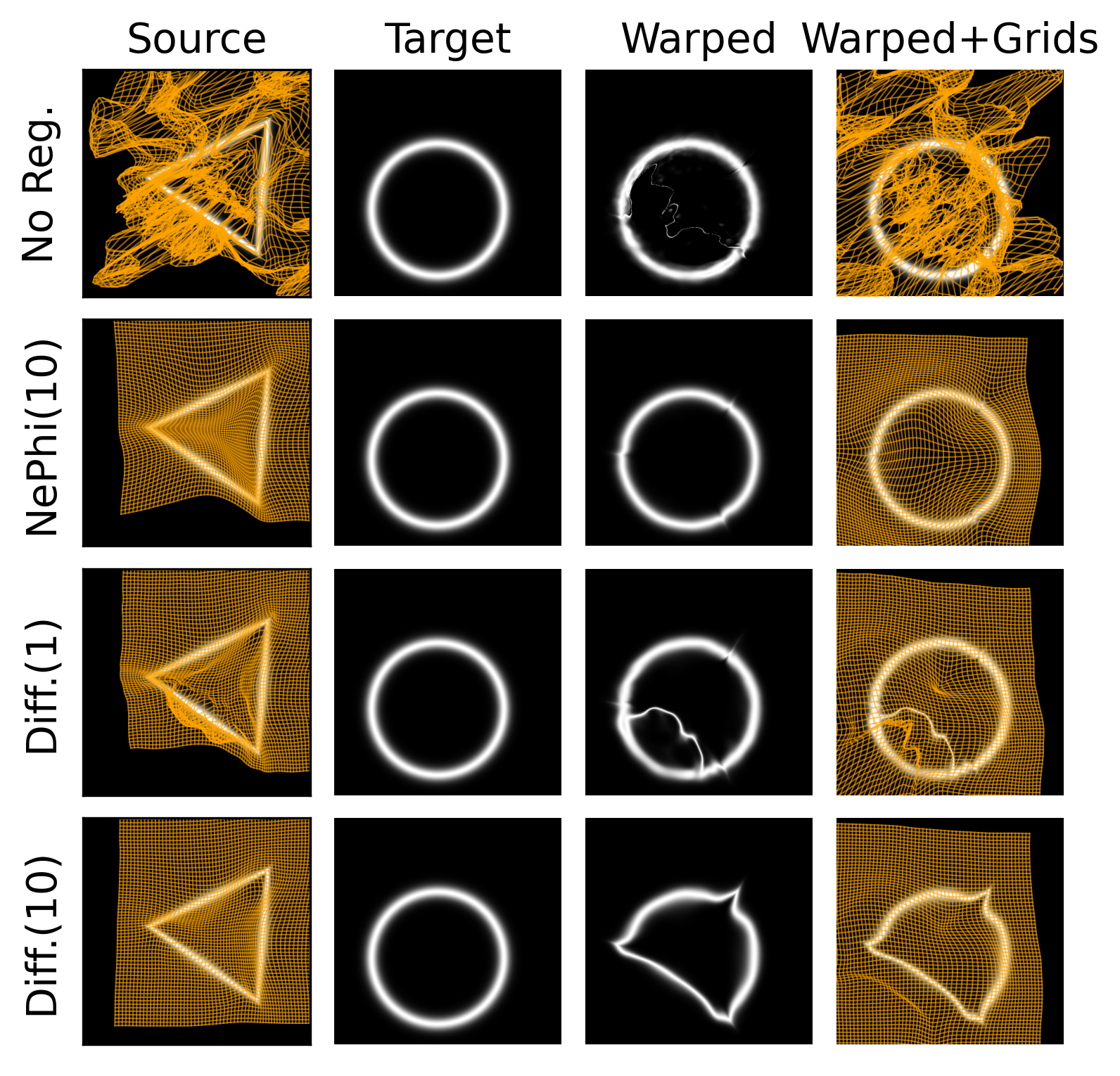}
        \caption{}
        \label{fig:nephi_vs_diffusion_example_hollow}
    \end{subfigure}
    \caption{Ablation study. (a) Performance of \texttt{NePhi} trained with various regularizer ratios, indicating that \texttt{NePhi}'s performance is insensitive to the regularizer's sampling ratio. (b) and (c) show the registration accuracy and regularity of the transformation field when using \texttt{NePhi} in the optimization-based setting (as in IO, but without predicting the latent code). The registration accuracy and regularity of the transformation field only deteriorates for regularizer ratios below $\sim 4^{-6}$. (d), (e) and (f) show comparisons between using the diffusion regularizer and gradient inverse consistency (\texttt{NePhi}) on NDF (Neural Deformation Field) transformations in the optimization setting.}
    \label{fig:nephi_vs_diffusion_opt}
\end{figure*}

\subsubsection{The number of points used for $\lreg$.}\label{sec:exp_ablation_number_of_sampling_points}
The underlying principle of \texttt{NePhi} that facilitates memory efficiency lies in its capability to preserve the regularity of deformation fields through the computation of regularity losses over a small number of sparse points, rather than densely across the entire image volume. In this section, we test this hypothesis and investigate the minimal set of points required to preserve regularity. We define the \textbf{regularizer ratio} $\%\lreg$ as the ratio between the number of sample points to evaluate the regularizer loss over the number of image voxels. To eliminate the influence of the similarity loss, we keep the similarity ratio fixed to 0.02 (seeSupp.~\ref{sec:implementation_details} for details) for all models in this section.

\textbf{Varying $\%\lreg$ in learning-based \texttt{NePhi}.} We train learning-based \texttt{NePhi} on the \textbf{T\&C} solid dataset. Fig.~\ref{fig:nephi_learning_varying_reg_ratio_loss} shows the Dice overlap scores between the fixed and the registered moving image. The number of folds ($\%|J|_{<0}$; not shown for this experiment) remained consistently close to zero for all models when varying $\%\lreg$. Fig.~\ref{fig:nephi_learning_varying_reg_ratio_loss} reveals that, at convergence of model training, both registration accuracy and the regularity of predicted deformation fields are minimally affected by variations in the regularizer ratio. Consequently, memory consumption for the evaluation of the regularizer loss can be dramatically reduced without affecting overall registration accuracy and regularity.

\textbf{Varying $\%\lreg$ with \texttt{NePhi}(IO).} We also tested \texttt{NePhi}(IO) when varying $\%\lreg$. Learning-based \texttt{NePhi} solves the T\&C registration problem well, so performing instance optimization starting from a trained learning-based \texttt{NePhi} results in flat training curves, starting and staying at the optimum. To emphasize convergence differences, we instead initialize the weights of the MLP and the latent code randomly, and then perform instance optimization. We randomly pick one image pair from the $\textbf{T\&C}$ as an example. 
Fig.~\ref{fig:nephi_optimization_varying_reg_ratio_dice} and Fig.~\ref{fig:nephi_optimization_varying_reg_ratio_fold} show the Dice overlap and transformation regularity respectively. We observe that reducing $\%\lreg$ does not significantly impact the regularity of the obtained transformations until $\%\lreg=\frac{1}{4^6}\sim0.0244\%$ is reached, corresponding to approximately 64 points only ($\%\lreg\times 512\times 512$). Importantly, since lowering the regularizer ratio does not affect regularity over a wide range of regularizer ratios, registration accuracy remains unaffected for these settings as well. This experiment underscores that, similar to learning-based \texttt{NePhi}, in this IO setting, satisfactory accuracy can be achieved by evaluating regularity using very few sample points for the regularizer loss, allowing for dramatic reductions in memory use for the evaluation of the regularizer loss.

\subsubsection{The different choices of $\lreg$.}\label{sec:comparing_different_regularizers}
We opt for gradient inverse consistency \mn{(GradICON)} regularization instead of the more commonly used diffusion regularizer~\cite{BalakrishnanZSG19} for deformation vector fields (DVF). This choice is motivated by the excellent registration performance obtained using \mn{the GradICON} regularizer for voxel-based learning-based registration \mn{and its ability to provide approximately diffeomoprhic transformations when combined with a deep registration network}~\cite{tian2022gradicon}. \mn{However, gradient} inverse consistency does not work in the optimization-based setting for a voxel-represented transformation (shown in Supp~\ref{sec:voxel_vs_nephi_with_gradient_inverse_consistency}). However, it does work for \texttt{NePhi}, enabling optimization over the transformation field itself instead of the entire network in the IO phase. 
This immediately also raises the question if gradient inverse consistency results in better solutions comparing to diffusion regularization for NDFs for instance optimization. 
To answer this question, we compare \texttt{NePhi} in the IO setting: i) \texttt{NePhi} without using a regularizer, and ii) \texttt{NePhi} replacing gradient inverse consistency with diffusion regularization. As discussed in Sec.~\ref{sec:exp_ablation_number_of_sampling_points}, to emphasize the convergence differences, we initialize \texttt{NePhi} (including latent codes and MLPs) from a random initialization. We present the result on one randomly generated pair of images from the \textbf{T\&C} hollow dataset (more examples in Supp.~\ref{sec:comparison_to_other_regularizer_continue}). Experimental results are shown in Fig.~\ref{fig:nephi_vs_diffusion_dice_hollow}, Fig.~\ref{fig:nephi_vs_diffusion_fold_hollow} and Fig.~\ref{fig:nephi_vs_diffusion_example_hollow}. Fig.~\ref{fig:nephi_vs_diffusion_fold_hollow} reveals that \texttt{NePhi} is the only approach that results in zero foldings during optimization. Although the number of folds ($\%|J|_{<0}$) also goes to zero for the models using the diffusion regularizer, the Dice results at convergence are lower compared to \texttt{NePhi} in Fig.~\ref{fig:nephi_vs_diffusion_dice_hollow}. 
Fig.~\ref{fig:nephi_vs_diffusion_dice_hollow} shows visualizations of the final registration results illustrating the good regularity properties of \texttt{NePhi} based on gradient inverse consistency regularization.

\section{Limitations}
Our method provides memory benefits during training and testing and allows graceful tradeoff between accuracy and runtime. However, it has the following limitations. First, the convergence is relatively slow. \texttt{NePhi} requires more training epochs (Supp.~\ref{sec:implementation_details}) and more iterations (800) during instance optimization to converge to a similar performance level as voxel-based registration neural networks (50 IO iterations are sufficient). This might be addressable by sampling more points to evaluate the losses which would increase memory requirements. Second, the current multi-resolution \texttt{NePhi} is still based on an initial voxel-based CNN registration model. For very high-resolution registration, such low resolution voxel-based networks may themselve consume a significant memory.

\section{Conclusion}
In this work, we introduced \texttt{NePhi}, which provides comparable registration accuracy, better deformation regularity, flexible settings to balance between test runtime and registration accuracy, and, importantly requires significantly less memory during training and instance optimization, compared to the current SOTA learning-based registration methods. \texttt{NePhi} opens up a large design space for learning-based registration (see Fig.~\ref{fig:lung_registration_methods_radar}) allowing a user to pick the best suited \texttt{NePhi} model for a particular registration task. \mn{In this work we demonstrated the good performance of \texttt{NePhi} within this design space on two challenging 3D image registration tasks: lung registration based on the compute tomography images of the COPDGene dataset and brain registration based on the magnetic resonance images of the HCP dataset.} We hope this work will inspire further explorations of generalizable neural deformation fields in learning-based registration methods and that it will open avenues for further research into high-resolution registration.

\section*{Acknowledgements}
The research reported in this publication was supported by the National Institutes of Health (NIH) under award numbers NIH 1R01HL149877, 1R01EB028283, \\1R01AR082684, 1R21MH132982, RF1MH126732, and 1R21EB035832. The content is solely the responsibility of the authors and does not necessarily represent the official views of the NIH. The brain imaging data were provided by the Human Connectome Project, WU-Minn Consortium (Principal Investigators: David Van Essen and Kamil Ugurbil; 1U54MH091657) funded by the 16 NIH Institutes and Centers that support the NIH Blueprint for Neuroscience Research; and by the McDonnell Center for Systems Neuroscience at Washington University. The lung imaging data was provided by the COPDGene study. 

\newpage

%
%
\bibliographystyle{splncs04}
\bibliography{references}

\newpage
\appendix
\clearpage

\section{Implementation Details}\label{sec:implementation_details}
\subsection{The structure of NePhi NDFs}
NePhi NDFs contain one forward transformation $\varphi_{\theta_1}$ and one backward transformation $\varphi_{\theta_2}$. The transformation maps are modeled by adding a displacement vector field $u$ to the identity transform, namely $\varphi_{\{\theta_1, \theta_2\}}(x, z(x))=x+u_{\{\theta_1, \theta_2\}}(x,z(x))$. To be noted, our proposed NDFs do not restrict the type of implicit function we use for $u(x, z(x))$, as long as the input is kept as $(x, z(x))$ and the output is a vector of the same dimension as $x$. This work uses Modulated Periodic Activations from Mehta et al.~\cite{mehta2021modulated} with MLPs as $u_\theta$.  $u_{\theta_1}$ and $u_{\theta_2}$ share the same structure but have separate parameters $\theta_1$ and $\theta_2$. We use two hidden layers with 512 hidden feature dimensions with each followed by sin activation functions~\cite{sitzmann2020implicit}. We use a Fourier feature mapping~\cite{tancik2020fourier} to embed the coordinates.

\subsection{The structure of CNN Encoder $f_{\theta_3}$}\label{sec:CNN_encoder_design}

For the encoder, we use a convolutional neural network (CNN) that accepts the concatenation of $I^A$ and $I^B$ as input and outputs the global latent code $z_g$. The CNN contains several layers, with each layer composed of one convolution,  one batch normalization, and one LeakyRelu. The kernel size and stride of the convolutional layer are set to 3 and 2, respectively.  The numbers of the input channel is 2 for the first layer. We use 6 layers and the numbers of the output channel are set to [8, 16, 32, 64, 128, 256]. After the CNN layers, we flatten the feature map and pass it to two fully connected layers. For high-resolution images, the dimension of the feature maps before the fully connected layers could be large. In such a scenario, we iteratively append layers with an output channel dimension of 256 as the final layer, provided that the dimension of the feature map remains greater than 3 after the last layer. 

\begin{figure*}[htp]
    \centering
    \includegraphics[width=\textwidth, trim={0 8cm 0 0cm}, clip]{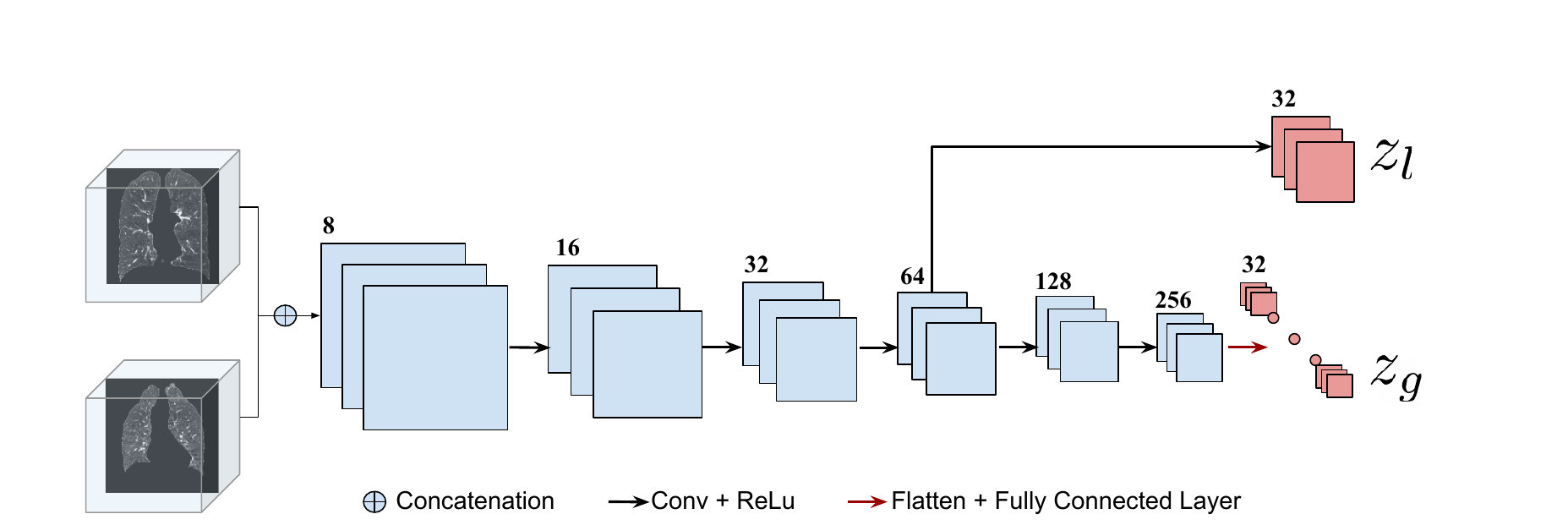}
    \caption{The structure of $f_{\theta_3}$ used in the experiments. The number of layers and the dimensions of the feature maps are kept the same for 2D and 3D registration. The position of the $z_l$ branch is adjusted according to the shape of the input image for each registration task to obtain a $z_l$ with a shape close to $16\times 16$ or $16\times 16\times 16$ for 2D and 3D registration respectively. }
    \label{fig:diagram_nephi_encoder}
\end{figure*}

We diverge from an intermediate layer, which is succeeded by a convolutional layer, to generate the local latent code $z_l$. The branching point is determined by the dimension of the intermediate feature map. We opt to branch from the feature map that most closely resembles $16\times16$ for 2D and $16\times 16\times 16$ for 3D feature maps. Fig.~\ref{fig:diagram_nephi_encoder} shows the structure of $f_{\theta_3}$ when the input image has the shape of $175\times{175}\times{175}$.

\subsection{Experimental Details}
We set the parameters for all experiments with $\lambda_1=10$, $\lambda_0=1e-2$, and dimensions for both $z_g$ and $z_l$ to 32. The similarity ratio is defined as $2e-2$, and the number of sampling points is constrained to a maximum of $10e5$. The regularizer ratio is established at $1e-4$, and the number of sampling points is restricted within the range of 10 to 500.

For the \textbf{COPDGene} dataset, the training of \texttt{NePhi} spans 2600 epochs, utilizing a batch size of 32. The model selected for evaluation is determined by the highest validation score on the validation set, where the validation score is assessed using the DICE score of the lung segmentation map. This configuration remains consistent across all experiments conducted on the \textbf{COPDGene} dataset. The networks are trained on either 4 RTX 3090 GPUs or 2 RTX A6000 GPUs.

In the inference phase, we measure the time period from passing the image pairs to the network to the network outputing the predicted dense DVF as the run-time. We compute the displacement vector at each voxel by batch with the batch size of 10000 for \texttt{NePhi} NDF.

\section{More Experimental Results}
\subsection{Registration Performance on DIRLab (Continued)}\label{sec:lung_registration_continue}
\begin{figure}[htp]
    \centering
    \includegraphics[width=\linewidth,, trim={0 0.3cm 0 0cm}, clip]{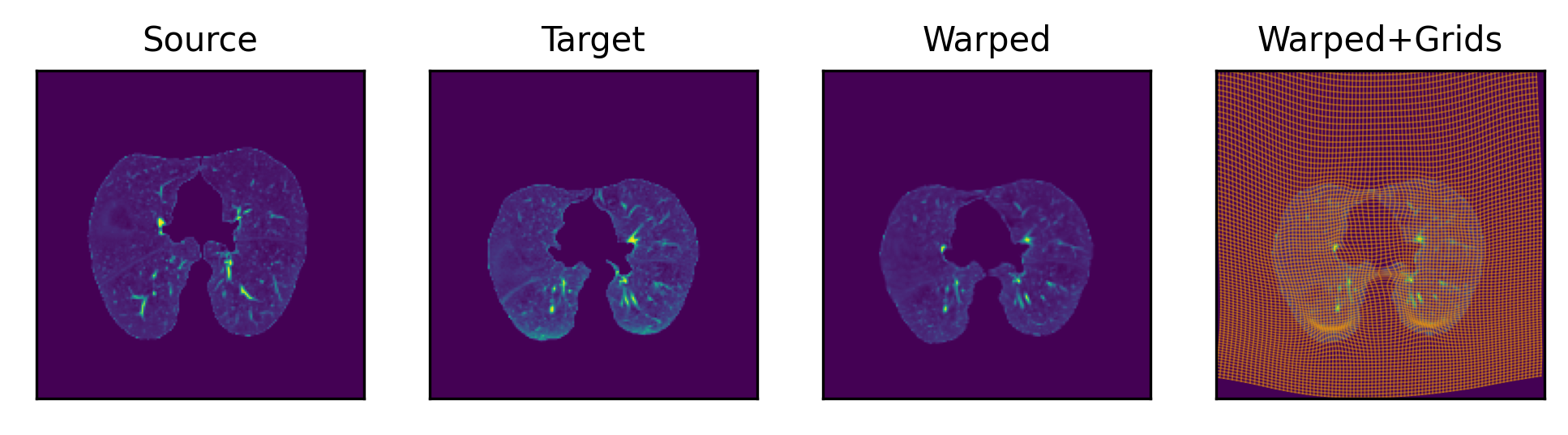}
    \caption{The registration result of case copd1 from DirLab dataset with the multi-resolution \texttt{NePhi} w/o IO.}
    \label{fig:lung_demo_supp}
\end{figure}

\begin{figure}[htp]
    \centering
    \includegraphics[width=\linewidth,, trim={0 0.3cm 0 0cm}, clip]{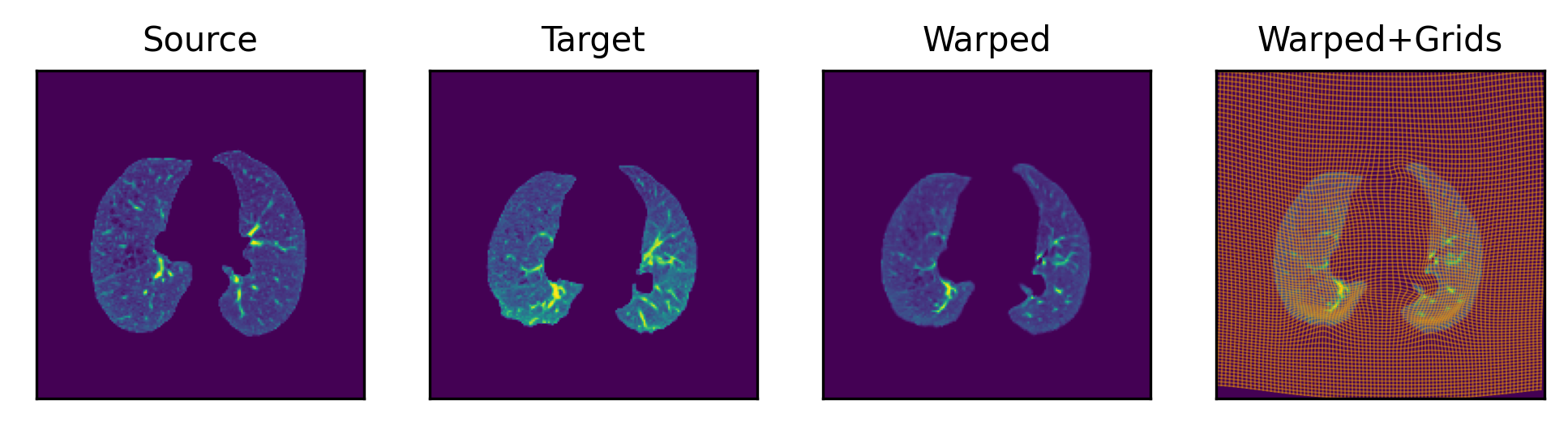}
    \caption{The registration result of case copd2 from DirLab dataset with the multi-res \texttt{NePhi} w/o IO. This example shows a scenario where the multi-resolution \texttt{NePhi} performs the worst out of the ten pairs of DirLab images.}
    \label{fig:lung_failure_demo_supp}
\end{figure}

To better visualize the performance of the methods we compare to and \texttt{NePhi} (Tab.~\ref{tab:registration_lung}) across four design considerations: registration accuracy, deformation regularity, peak memory consumption during training, and inference run-time, we linearly normalize the values to [0,1] and show the visualization in Fig.~\ref{fig:lung_registration_methods_radar}. The range we used to normalize the mTRE, $\%|J|_{<0}$, inference run-time, and memory consumption are $0mm \sim 5mm$, $0 \sim3e-2$, $0s\sim 70s$, and $0M \sim 1.5e4M$, respectively. Fig.~\ref{fig:lung_demo_supp} shows an example Dirlab registration result for \texttt{NePhi}. In addition, Fig.~\ref{fig:lung_failure_demo_supp} shows the example where \texttt{NePhi} performs the worst out of the DirLab dataset.

\subsection{Regularization Analysis between Voxel- and Implicit-representation}\label{sec:voxel_vs_nephi_with_gradient_inverse_consistency}
\begin{figure*}[htp]
    \centering
    \begin{subfigure}[t]{0.33\textwidth}
        \centering
        \includegraphics[width=\textwidth]{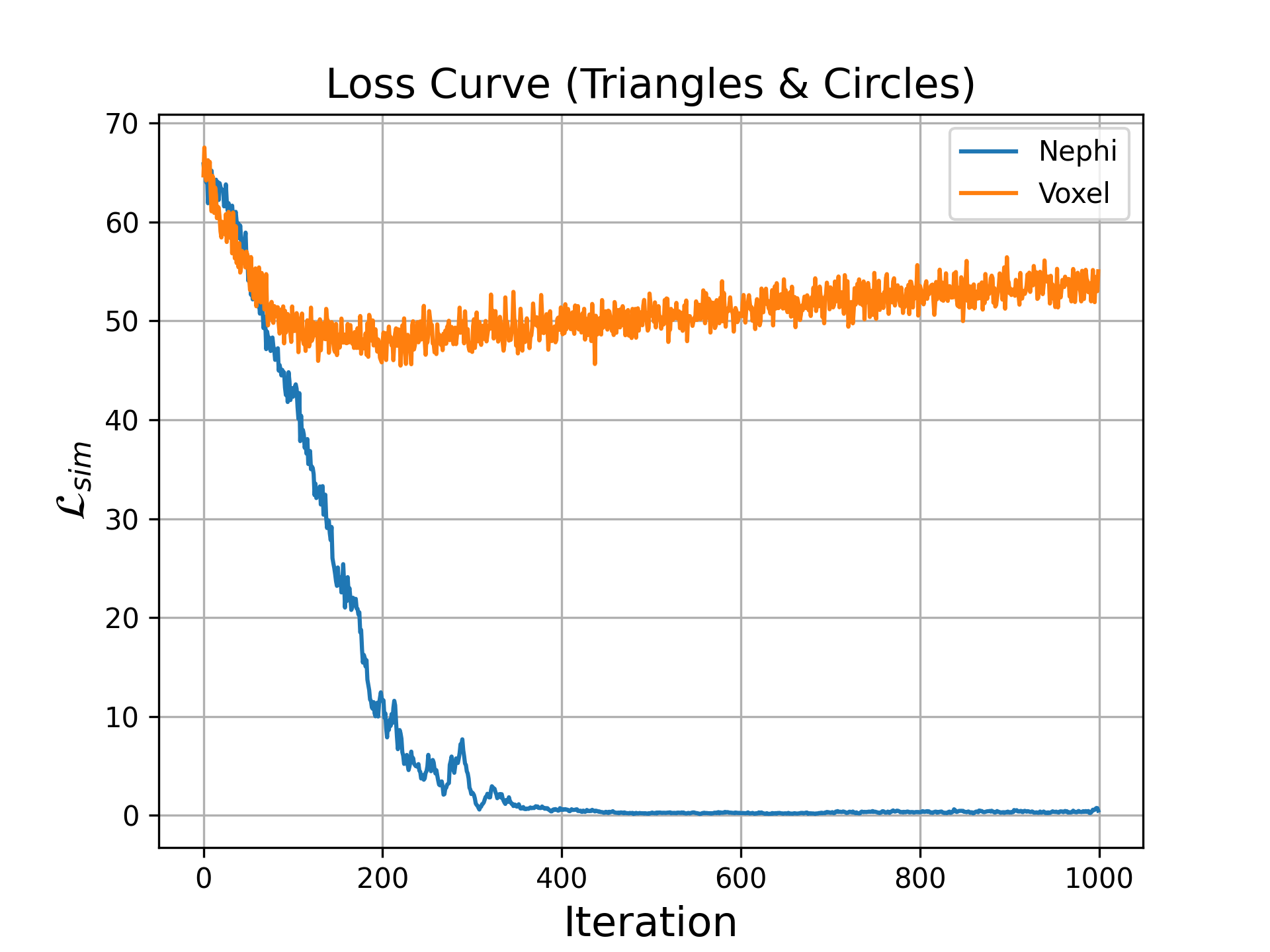}
        \caption{$\mathcal{L}_{sim}$ of using \texttt{NePhi} NDFs and voxel-transformation in instance optimization.}
        \label{fig:nephi_vs_voxel_opt_loss}
    \end{subfigure}
    \begin{subfigure}[t]{0.33\textwidth}
        \centering
        \includegraphics[width=\textwidth]{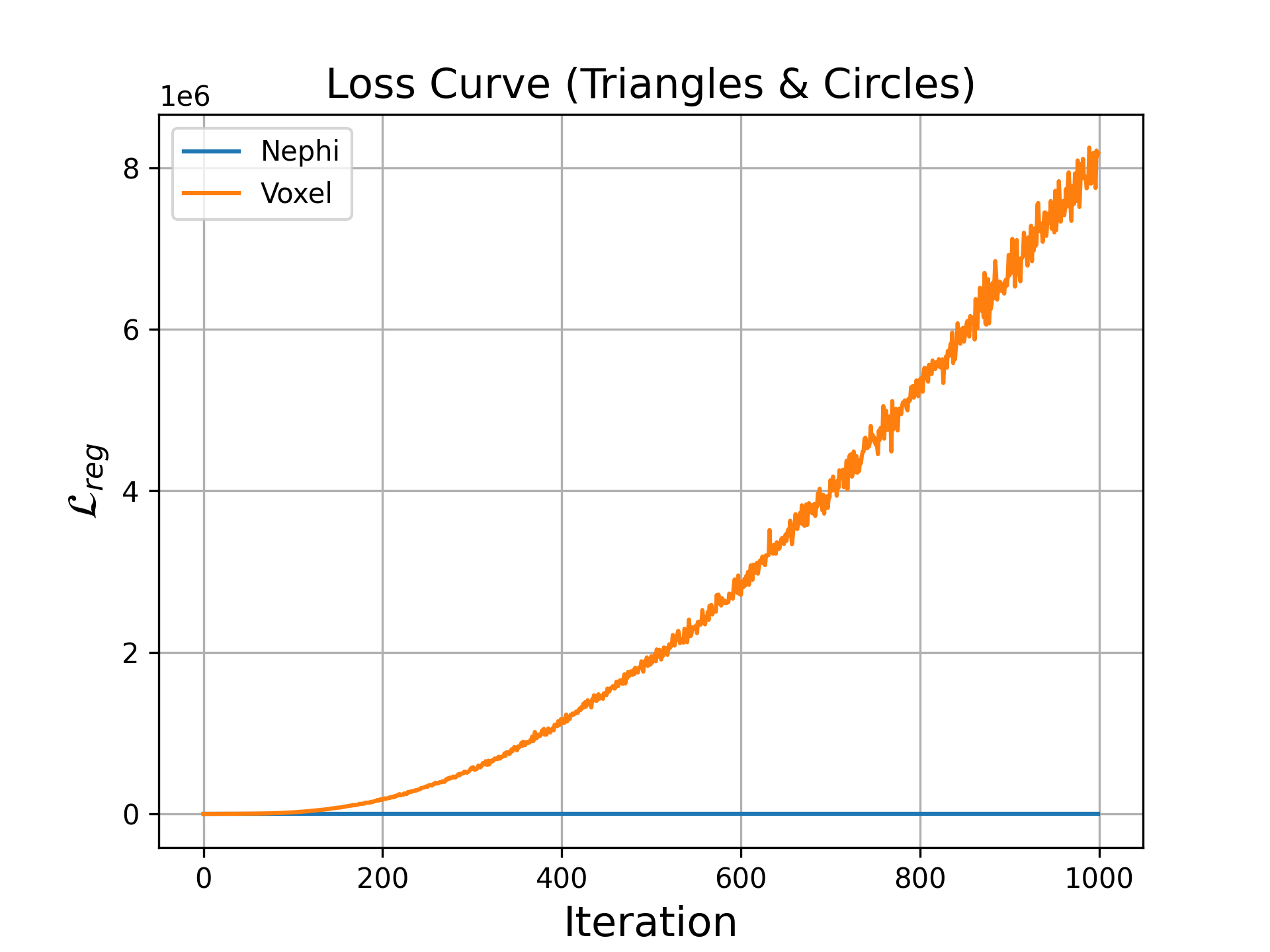}
        \caption{$\mathcal{L}_{reg}$ curve of \texttt{NePhi} NDFs and voxel-transformation in instance optimization.}
        \label{fig:nephi_vs_voxel_opt_reg}
    \end{subfigure}
    \caption{Experiments investigating the regularity properties of gradient inverse consistency show that gradient inverse consistency does not function in the optimization-based registration framework when combined with voxel-represented transformation (\textcolor{orange}{Orange curve}). However, it does work when combined with \texttt{NePhi} NDFs (\textcolor{NavyBlue}{Blue curve}).}
\end{figure*}

To pursue optimal registration accuracy in practical applications, the adoption of a post-processing step called instance optimization (IO) is common following the use of a learning-based registration network for inference. This involves conducting optimization over either the neural network or the inferred transformation map for a single pair of images, aiming for further precision. However, optimizing over voxel-represented transformations with gradient inverse consistency may yield a non-regular solution (see Fig.~\ref{fig:nephi_vs_voxel_opt_reg}). In this scenario, one must optimize the parameters of the overall CNN instead of the inferred transformation map in IO (as done in ~\cite{tian2022gradicon}), incurring a significant memory cost during testing.

Given that \texttt{NePhi} NDFs contain MLPs that likely possess implicit regularization, it is reasonable to anticipate that \texttt{NePhi} NDFs alone (without $f_{\theta_3}$) can achieve regularity. To test this hypothesis, we compared the loss curves of \texttt{NePhi} NDFs and voxel-based transformation with gradient inverse consistency in Fig.~\ref{fig:nephi_vs_voxel_opt_loss} and Fig.~\ref{fig:nephi_vs_voxel_opt_reg}. In our proposed framework, the NDFs and voxel-based transformation are initialized by the inferred transformation of a trained registration network. In this experiment, to emphasize the convergence difference, we initialize the NDFs and voxel-based transformation from an identity transform. Notably, \texttt{NePhi} NDFs yield better deformation regularity than voxel-based transformation as shown in Fig.~\ref{fig:nephi_vs_voxel_opt_reg}. Moreover, the registration using voxel-based transformation converges to a much poorer similarity measure compared to \texttt{NePhi} NDFs in Fig.~\ref{fig:nephi_vs_voxel_opt_loss}.

\subsection{Comparing Different Regularizers (continue)}\label{sec:comparison_to_other_regularizer_continue} 
\begin{figure*}[htp]
    \centering
    \begin{subfigure}[t]{0.34\textwidth}
        \centering
        \includegraphics[width=\textwidth]{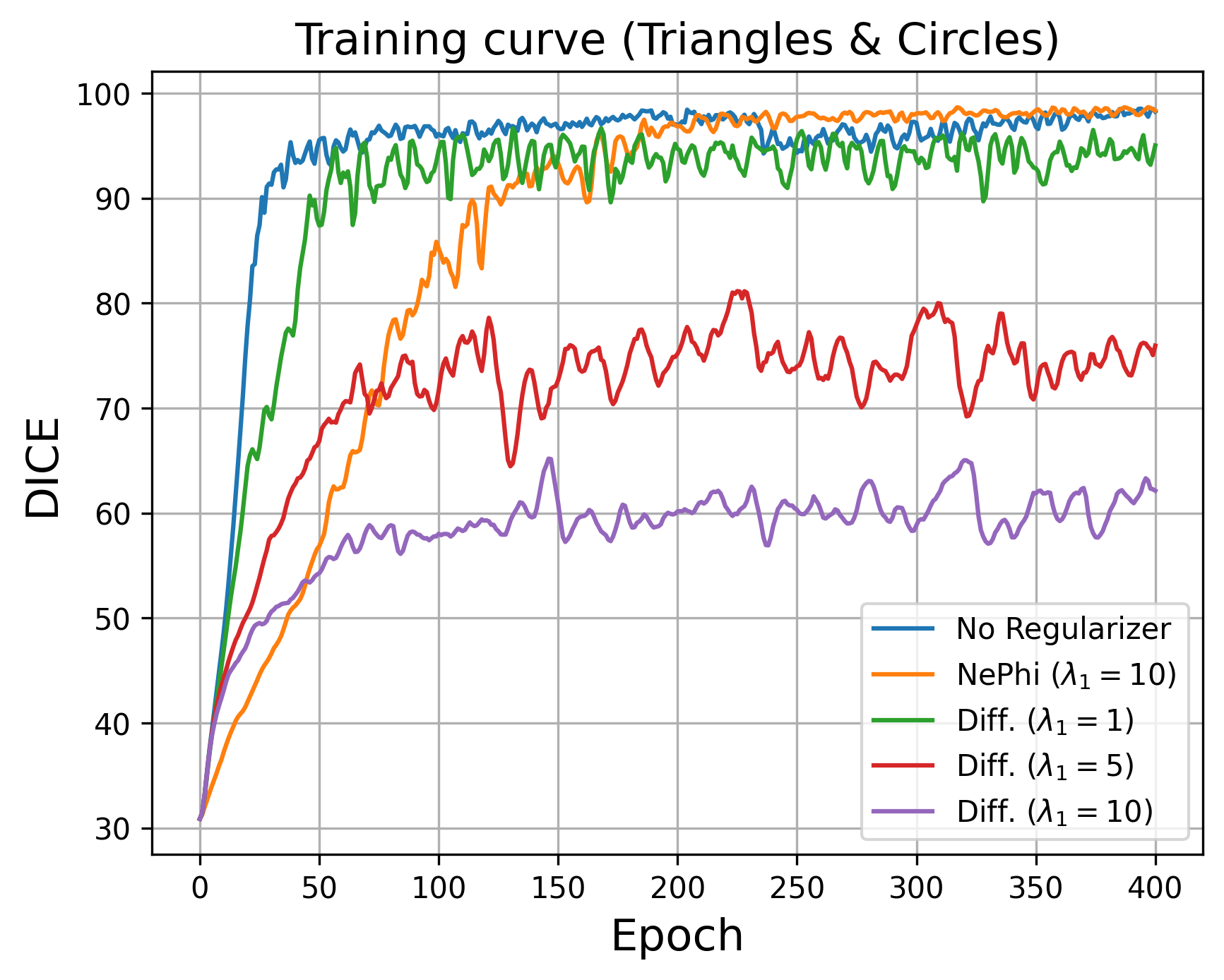}
        \caption{}
        \label{fig:nephi_vs_diffusion_dice_solid}
    \end{subfigure}
    \begin{subfigure}[t]{0.34\textwidth}
        \centering
        \includegraphics[width=\textwidth]{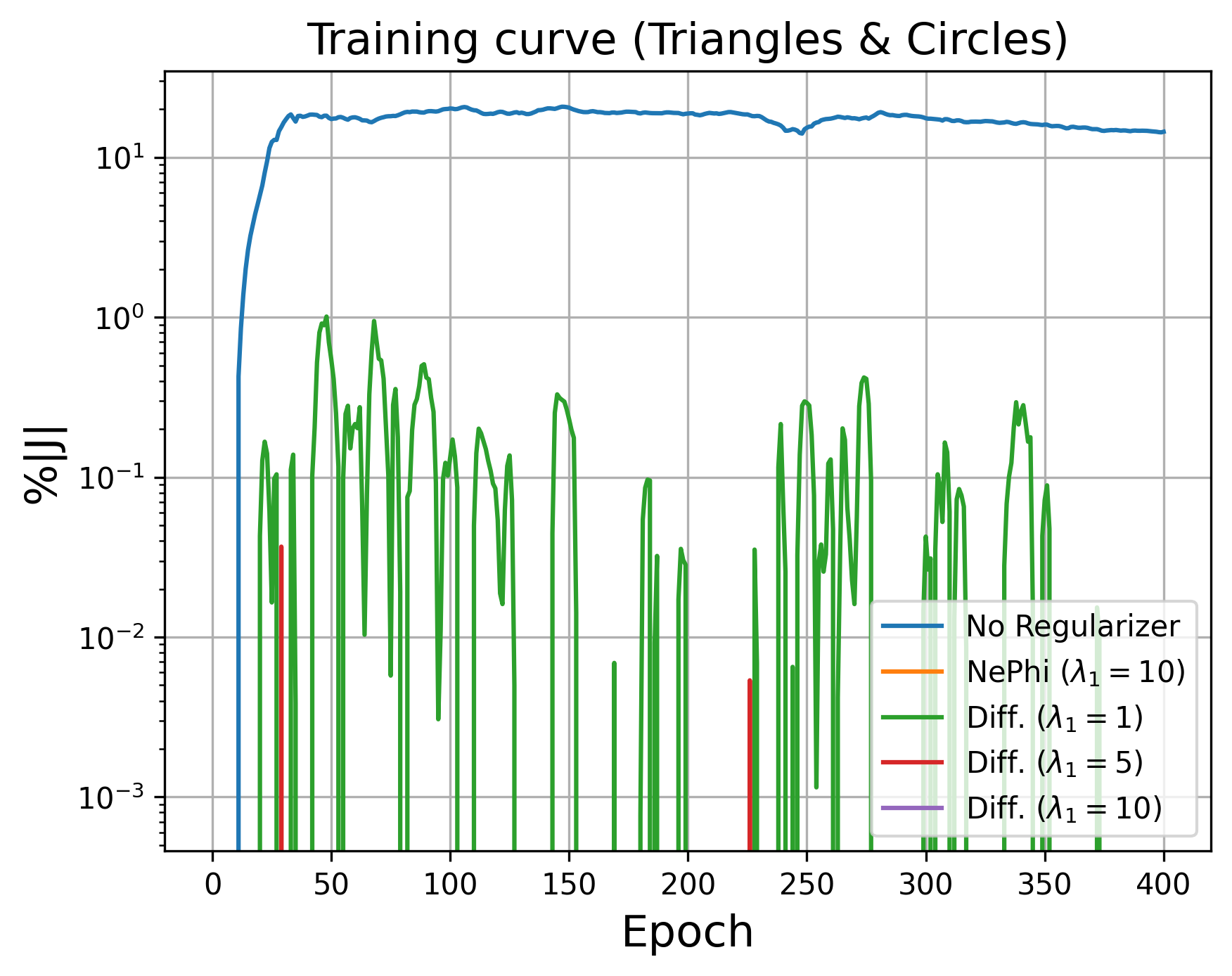}
        \caption{}
        \label{fig:nephi_vs_diffusion_fold_solid}
    \end{subfigure}
    \begin{subfigure}[t]{0.28\textwidth}
        \centering
        \includegraphics[width=\textwidth]{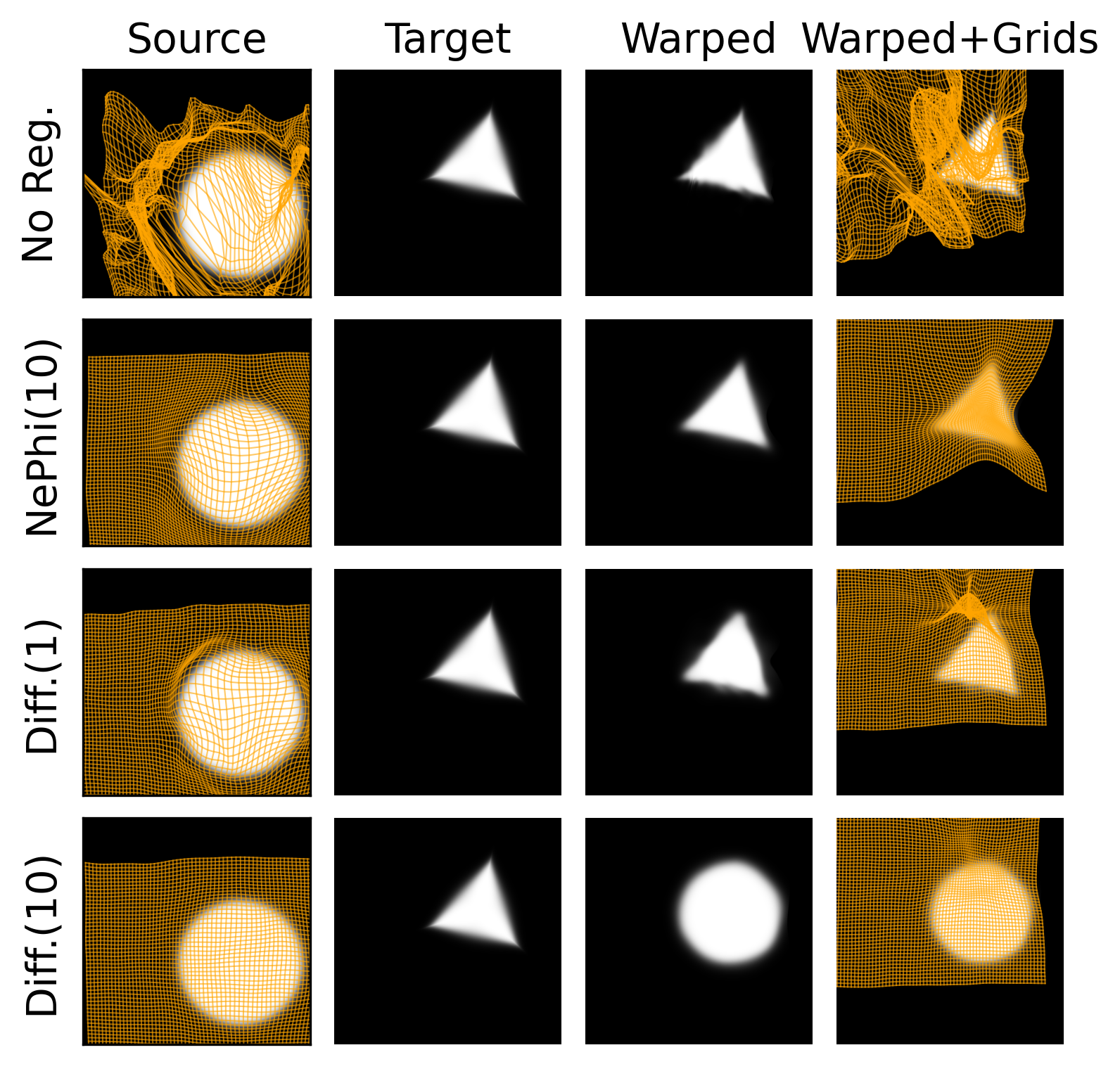}
        \caption{}
        \label{fig:nephi_vs_diffusion_example_solid}
    \end{subfigure}
    \caption{Comparisons between diffusion regularizer and gradient inverse consistency (\texttt{NePhi}) with NDFs in instance optimization. (a), (b) and (c) are the corresponding DICE curve, $\%|J|_{<0}$ curve and qualitative results of one pair of \textbf{C\&T} solid images.}
    \label{fig:nephi_vs_diffusion_opt_continue}
\end{figure*}

In Sec.~\ref{sec:comparing_different_regularizers}, we have compared gradient inverse consistency regularizer with other regularizers on \textbf{T\&C} hollow dataset. In Fig.~\ref{fig:nephi_vs_diffusion_opt_continue}, we show the results on \textbf{T\&C} solid dataset with the same experiment setting as in sec.~\ref{sec:comparing_different_regularizers}.

\begin{figure}[htp]
    \centering
    \includegraphics[width=0.5\linewidth]{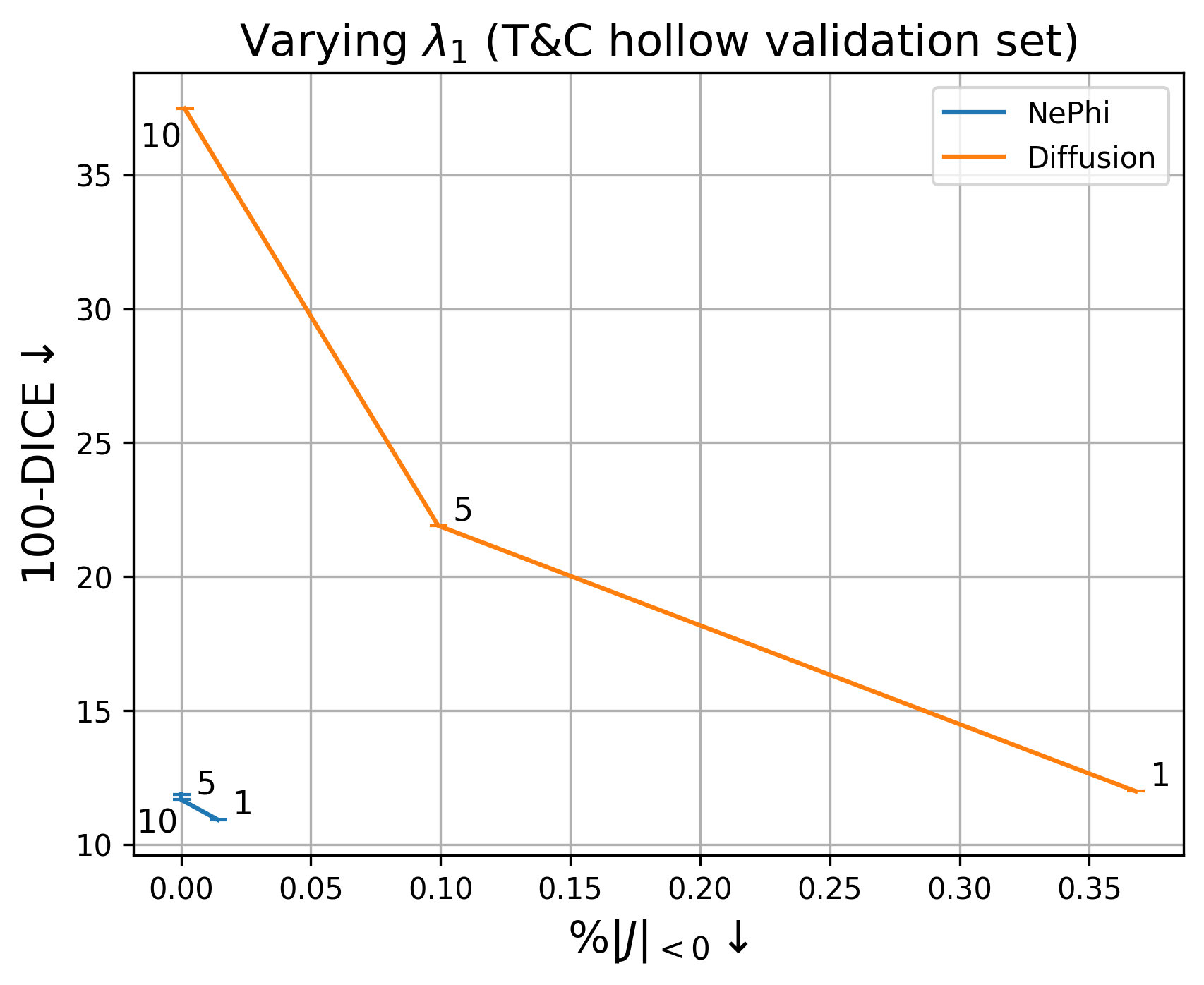}
    \caption{The similarity and regularity trade-off comparison between gradient inverse consistency and diffusion regularizer with \texttt{NePhi} NDFs.}
    \label{fig:nephi_vs_diffusion_varying_lamb}
\end{figure}

In \texttt{GradICON}~\cite{tian2022gradicon}, the authors conducted an experiment to compare the trade-off of the similarity and regularity between different regularizers when used with a voxel-representation for the DVF. We also conduct the same experiment on \textbf{T\&C} hollow to show the trade-offs between gradient inverse consistency and diffusion regularizer with \texttt{NePhi} NDFs. Fig~\ref{fig:nephi_vs_diffusion_varying_lamb} shows the results.

\subsection{Interpolation of the Learned Latent Space} 
\begin{figure*}[htp]
\centering
    \centering
    \includegraphics[width=1\linewidth]{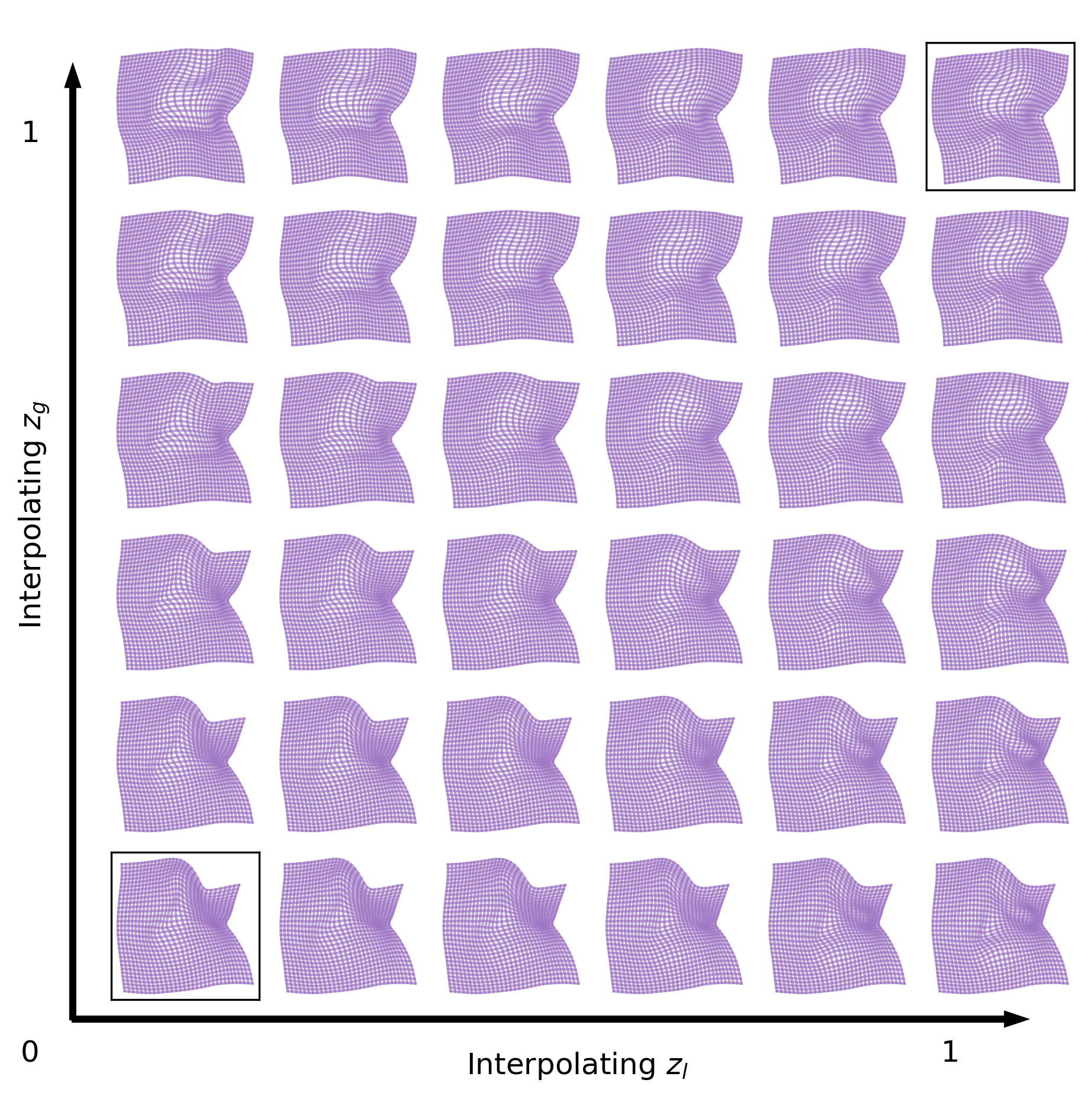}
    \caption{Interpolation of two latent codes \emph{predicted} by \texttt{NePhi} that are trained on \textbf{T\&C} hollow dataset.}
    \label{fig:latent_interpolation}
\end{figure*}

The expressiveness of the latent space of \texttt{NePhi} is critical, given that we want \texttt{NePhi} to be used in a learning-based setting. We conduct the following experiment to study the latent space learned by NePhi. We train $f_{\theta_3}$ with \texttt{NePhi} on the \textbf{T\&C} hollow dataset. Subsequently, we perform interpolation between the latent codes predicted by $f_{\theta_3}$ using two randomly selected image pairs from the training set. From Fig.~\ref{fig:latent_interpolation} we observe a smooth transition of transformations from the bottom left to the top right. Additionally, it is noteworthy that as the interpolation ratio of $z_g$ changes from 0 to 1, the global deformation becomes increasingly similar to the transformation in the top right. This finding suggests that $z_g$ and $z_l$ capture deformations at different levels, which aligns with our intended design.

\end{document}